\documentclass[runningheads]{llncs}
\usepackage{graphicx}
\usepackage{comment}
\usepackage{amsmath,amssymb} 
\usepackage{color}
\usepackage{booktabs}       
\usepackage{amsmath}
\usepackage{subfig}
\usepackage{caption}
\usepackage{float}
\DeclareMathOperator*{\argmax}{arg\,max} 
\newcommand*{\affmark}[1][*]{\textsuperscript{#1}}

\begin{document}
\pagestyle{headings}
\mainmatter
\def\ECCVSubNumber{3368}  

\title{Facial Landmark Correlation Analysis} 

\titlerunning{Facial Landmark Correlation Analysis}
%
\author{First Author\inst{1}\orcidID{0000-1111-2222-3333} \and
Second Author\inst{2,3}\orcidID{1111-2222-3333-4444} \and
Third Author\inst{3}\orcidID{2222--3333-4444-5555}}

\author{
Yongzhe Yan\affmark[1] \ \ \ \ Stefan Duffner\affmark[2]\ \ \ \ Priyanka Phutane\affmark[1]\ \ \ \ Anthony Berthelier\affmark[1]\ \ \ \ Christophe Blanc\affmark[1]\ \ \ \
Christophe Garcia\affmark[2]\ \ \ \ Thierry Chateau\affmark[1]
}

\authorrunning{Y. Yan et al.}
%
\institute{Universit\'e Clermont Auvergne, CNRS, SIGMA, Institut Pascal, Clermont-Ferrand, France \and
Universit\'e de Lyon, CNRS, INSA-Lyon, LIRIS, UMR5205, France 
\email{yongzhe.yan@etu.uca.fr}}
\maketitle

\begin{abstract}
We present a facial landmark position correlation analysis as well as its applications. 
Although numerous facial landmark detection methods have been presented in the literature, few of them explicitly take into account the inherent relationship among landmarks. 
To reveal and interpret this relationship, we propose to analyze 
landmark correlation by using Canonical Correlation Analysis~(CCA). 
We experimentally show that the dense facial landmark annotations in current benchmarks are strongly correlated.

We propose two applications based on this analysis. First, by analyzing the landmark correlation, we gain some interesting insights into the predictions of different landmark detection models (including random forests model and CNN models). We also demonstrate how CNNs progressively learn to predict facial landmarks.
Second, we propose a few-shot learning method that allows to considerably reduce the manual effort for dense landmark annotation.
Overall, this landmark correlation analysis provides new perspectives for the research on facial landmark detection.

\end{abstract}

\section{Introduction}

\begin{figure}[t]
\centering
\subfloat[300W Train Subset]{\includegraphics[width=0.447\columnwidth]{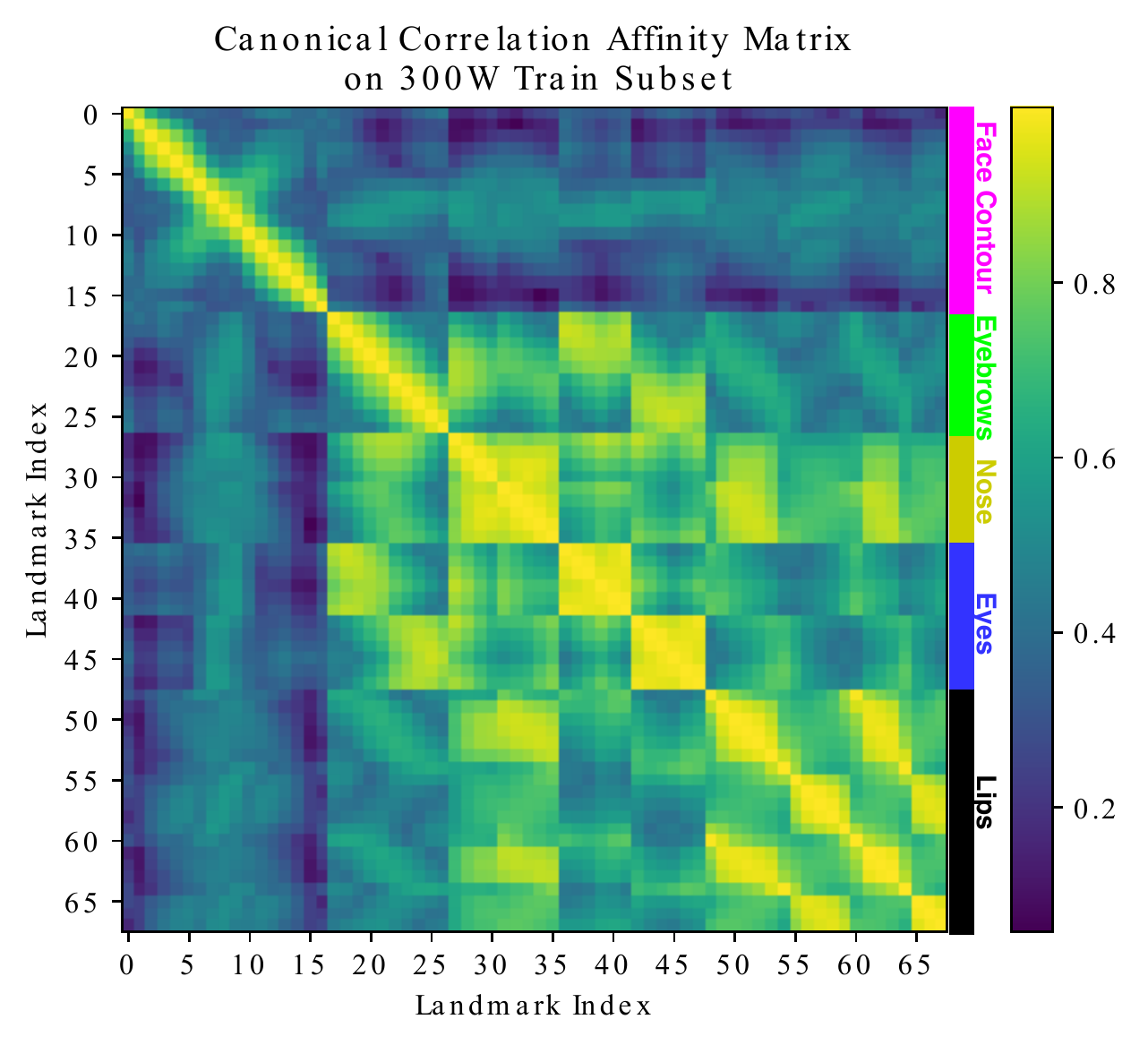}}
\subfloat[Landmark Index]{\includegraphics[width=0.43\columnwidth, height=4.96cm]{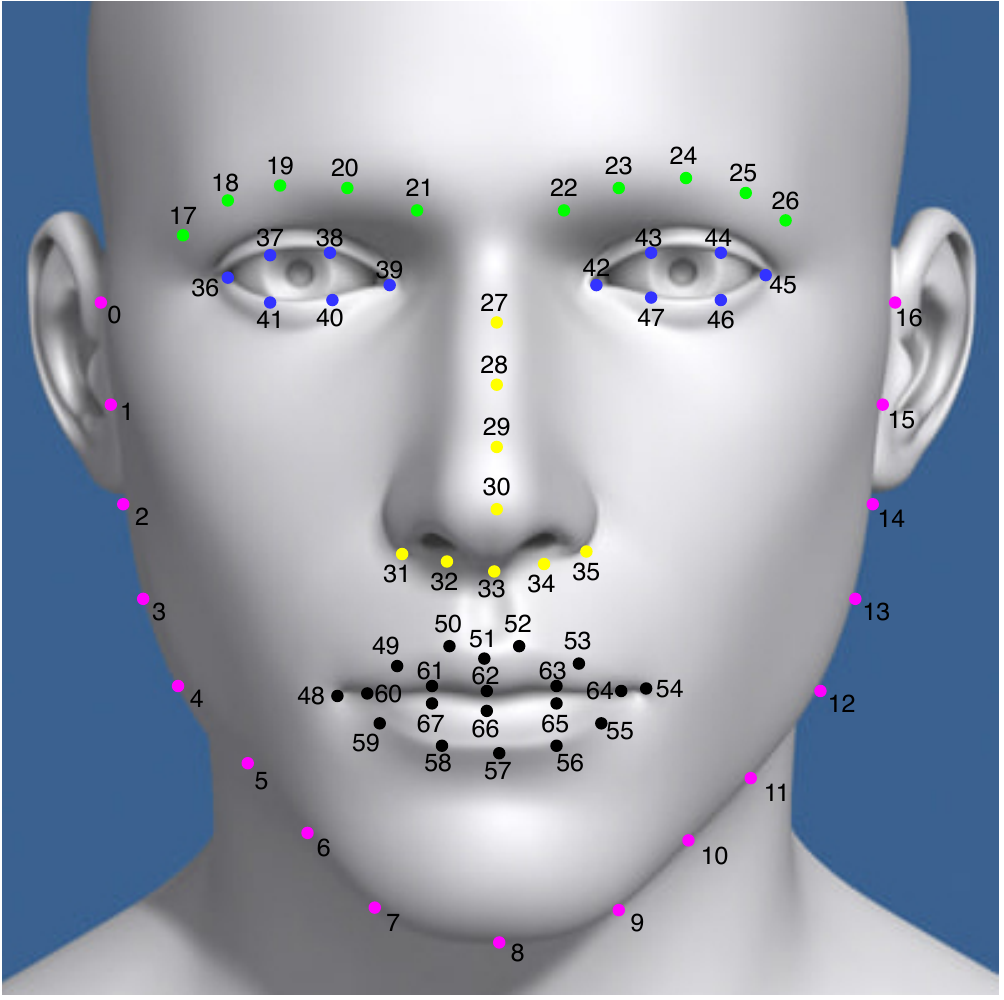}}
\caption{Facial landmark correlation analysis on the ground truth of 300W train subset. (a) Canonical Correlation Analysis (CCA) affinity matrix. Bright yellow colored points indicate that the two respective landmarks are highly correlated, and dark blue color indicates low correlation. (b) Illustration of the annotated landmark indices for the 300W dataset. Best viewed in color.}
\label{fig:teaser} 
\end{figure} 

Facial landmark detection is an active research topic of computer vision in recent years. The aim is to retrieve the coordinates of a given number of fiducial points on a face image. 
It is an important prerequisite for numerous applications such as face recognition~\cite{Ding2016}, 3D face reconstruction~\cite{jackson2017vrn}, facial expression analysis~\cite{martinez2017automatic}. 
Most of the facial landmarks are positioned on the face contour, eyebrows, eyes, nose and lips.
\par

 \textbf{What is facial landmark correlation?} Due to the shape and motion of real 3D objects, there exists a natural correlation between landmarks positioned on these objects (e.g.~faces, human body, hands or other objects), also in corresponding 2D projections.
 Especially for faces, the correlation among landmarks is very strong due to the following two reasons: First, the human face is more rigid than the entire human body or hands, which have more articulations and may be observed from any point of view under severe rotations or deformations. 
 Second, recently-released facial landmark datasets are densely annotated with up to 98 landmarks~\cite{wu2018look}, exhibiting an even stronger correlation. Therefore, in this paper, we focus on the correlation of dense facial landmark annotation in 300W dataset~\cite{sagonas2013w300} (68 landmarks, see Fig.~\ref{fig:teaser}) and WFLW dataset~\cite{wu2018look} (98 landmarks). 
\par

\textbf{Motivation of this analysis:}
The standard evaluation metric for facial landmark detection is the Normalized Mean Error (NME). 
NME is the averaged Euclidean distance between each predicted landmark and ground truth, normalized by the inter-ocular distance.
A smaller NME indicates a more precise prediction and vice-versa.
Other commonly used metrics, including Failure Rate (FR), Cumulative Error Distribution (CED), and Area Under Curve (AUC), are all based on NME. 
\par

However, we think that the NME can not describe all aspects of the model prediction.
A large NME can signify that the prediction is not precise, but it can not reflect how the prediction is mistaken. We illustrate this limitation in the following example.
\par

Current deep learning-based state-of-the-art methods can be categorized into two types: Coordinate Regression CNNs (CR-CNNs) and Heatmap Regression CNNs (HR-CNNs)~\cite{yan2018survey,wu2017survey}.
CR-CNNs predict the numeric X and Y coordinate values of each landmark in the last Fully-Connected (FC) layer. 
HR-CNNs adopt Fully Convolutional Neural Network~\cite{long2015fully} architectures that estimate a spatial probability map for each landmark.
That is, the value of each pixel on the heatmap represents the presence probability of the landmark at this pixel~\cite{wei2016convolutional}.\par

\begin{figure}[t]
  \centering
  \includegraphics[width=0.85\textwidth]{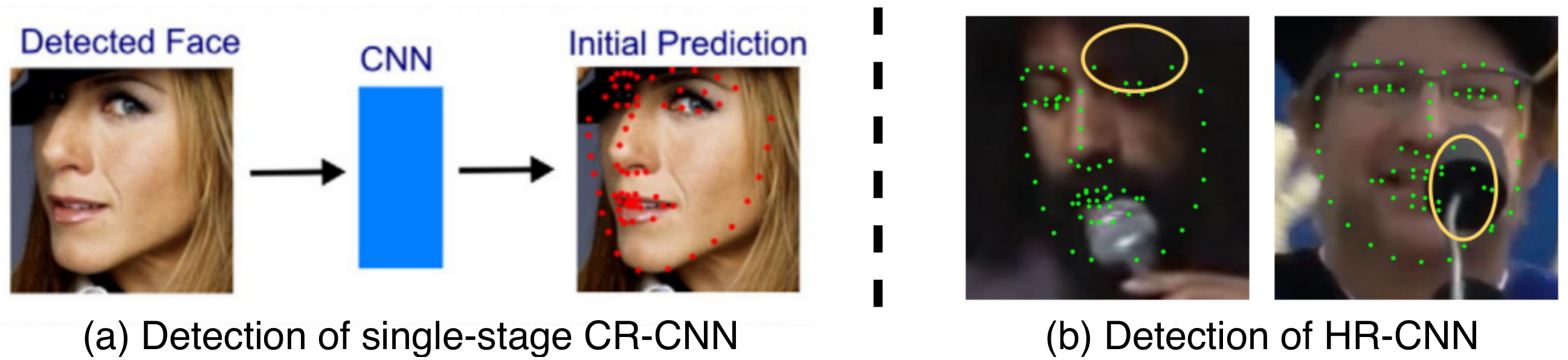}
  \caption{Illustration of the weaknesses of single-stage CR-CNNs and HR-CNNs. Figure (a) is taken from~\cite{fan2016approaching} and figure (b) is taken from~\cite{liu2019semantic}.}
  \label{fig:weakness}
\end{figure}

Each model has its strengths and weaknesses. HR-CNNs show a strong capability of handling complex pose variations. However they globally lack robustness, and in failure cases, landmarks are predicted at unreasonable positions which are far away from the ground truth (see Fig.~\ref{fig:weakness}~(b)). 
On the other hand, CR-CNNs are generally more efficient in terms of computation and memory usage, but also locally less precise. 
The prediction of single-stage CR-CNNs is usually constrained in a reasonable shape similar to the ground truth, being not extremely precise (see Fig.~\ref{fig:weakness}~(a)). 
This investigation can be confirmed by the current research trend. Most of the latest HR-CNNs aim at reinforcing the robustness of the detection by introducing global constraints~\cite{Valle_2018_ECCV,liu2019semantic,merget2018robust,zou2019learning} or temporal consistency~\cite{tai2018towards}. 
However, the recent CR-CNNs enhance local precision using coarse-to-fine frameworks~\cite{trigeorgis2016mnemonic,fan2016approaching,lv2017deep,chen2017delving,kowalski2017dan,he2017fully,feng2018wing}.\par

For instance, the models whose results are illustrated in Fig.~\ref{fig:weakness}~(a) and Fig.~\ref{fig:weakness}~(b) may have similar NME. Nevertheless, these two models have distinct characteristics.  
In this case, we think that landmark correlation can be the key to explain and, furthermore, to quantify the weaknesses of the two models.
We assume that the local imprecision problem of CR-CNNs is due to the fact that the predicted landmark positions are too much correlated (or regularized). 
In contrast, for HR-CNNs, the ``outliers'' predicted in unreasonable positions can be considered as a violation of the natural landmark correlation. \par

We want to make clear that the landmark correlation can not be used as a stand-alone evaluation metric, though it provides a new perspective to interpret the model prediction. Similar correlation compared to the ground truth is a necessary but not sufficient condition to precise prediction.  Even identical correlation does not ensure precise prediction. However, big correlation difference between prediction and ground truth can indicate imprecise prediction. 

Our contributions can be summarized as follows:
\begin{itemize}
\item[$\bullet$] We present a CCA-based correlation analysis as a novel tool to interpret and quantify the relationship among a set of landmarks (Sect.~\ref{sec:CCA}).\par
\item[$\bullet$] We use this model-agnostic correlation analysis to gain some insights on the three most popular facial landmark detection models in the last decade, including cascaded random forest, CR-CNNs and HR-CNNs (Sect.~\ref{sec:CNN_interpretation}).\par
\item[$\bullet$] With the help of landmark correlation, we propose a few-shot learning method to reduce the manual effort for dense landmark annotation (Sect.~\ref{sec:Few-shot Learning}). 
\end{itemize}

\section{Related Work}
\textbf{Facial landmark detection in the last decade:} In 2010, Doll\'ar et al. proposed Cascaded Pose Regression~\cite{dollar2010cascaded}, which laid the foundation for several well-known cascaded regression methods including SDM~\cite{xiong2013supervised}, ESR~\cite{cao2014face} and ERT~\cite{kazemi2014one}. 
In the deep learning era, cascaded CR-CNNs~\cite{sun2013deep,zhang2014coarse,trigeorgis2016mnemonic} continue to follow its general coarse-to-fine structure.
HR-CNNs~\cite{wei2016convolutional,newell2016stacked,Bulat_2017_ICCV}, originally introduced in 2005~\cite{duffner2005connexionist}, gained much popularity in recent years.
In this paper, we propose facial landmark correlation analysis to take a closer look into three of the most important models in the last decade: Cascaded Random Forest model~\cite{kazemi2014one}, CR-CNN~\cite{fan2016approaching} and HR-CNN~\cite{Bulat_2017_ICCV}.
\par

\textbf{Component analysis in facial landmark detection:}
The use of Principal Component Analysis (PCA), especially the 3DMM model~\cite{blanz1999morphable}, is of great importance in the current research of face analysis.
PCA has been used for facial landmark detection since 1995~\cite{cootes1995active} in the Point Distribution Model.
PCA is used to analyze the shape variance with respect to the mean shape, including face rotation, facial expressions and identity variance. 
The biggest difference between the PCA and our CCA study is that our CCA study analyzes the relationship between individual facial landmarks while PCA focuses on the global face shape. \par  
\par

\textbf{CNN Interpretation via CCA:}  Using CCA to interpret CNN representations is an emerging subject. In the recent work~\cite{svcca,morcos2018insights,kornblith2019similarity}, they used CCA to analyze the representations of \textit{different CNNs} and gained some insights on the learning process. 
However, as we will show in Sect.~\ref{sec:4.3_Network Learning Dynamics}, we use CCA to analyze the correlation between \textit{different neurons} in the same layer. 

\textbf{Few-shot learning for facial landmark detection:} Few-shot learning, or weakly supervised learning, is now attracting increasing attention in the community.
A recent work~\cite{dong2019teacher} proposed a mechanism to enable the training on fewer labeled images. In this paper, we focus on how to learn with fewer \textit{landmarks} rather than fewer \textit{images}. 
\par

We assume that a dense format can be transferred from a sparse format. 
This is not new and has already been proved in several work which focus on transferring the data between two annotations with different semantic meanings~\cite{smith2014collaborative,zhu2014transferring,zhang2015leveraging}.
We also find similar ideas in some existing coarse-to-fine approaches~\cite{lv2017deep,chen2017delving,shao2016learning,shao2018multicenter}, where the entire set of landmarks is divided into several partitions inside which the authors assume a strong correlation. 
Specifically, Tan~\textit{et~al.}~\cite{tan2017extra} proposed a few-shot learning method to reconstruct the global dense shape from a sparse landmark format.
DeCaFA~\cite{Dapogny_2019_ICCV} can be trained with sparsely annotated examples by exploiting landmark-wise attention.
However, in the above works, they mainly focused on how to improve the model performance given the pre-defined sparse format. The choice of the sparse format is heuristic. 

In contrast, in this paper, we focus on how to find the best sparse format that will most benefit the few-shot learning. 
Our selection of the sparse landmark format is entirely based on the statistics of the underlying data.
This idea is inspired by the work on multi-task learning~\cite{zamir2018taskonomy,Li2019Task}. In the context of facial landmark, we consider the prediction of each landmark as an individual task.\par

\section{Facial Landmark Correlation Analysis}
\label{sec:CCA}
\subsection{Canonical Correlation Analysis~\cite{hotelling1936relations}}
Given a $p$-dimensional random variable $\mathbf{U} \in \mathbb{R}^{p}$ and a $q$-dimensional variable $\mathbf{V} \in \mathbb{R}^{q}$, CCA aims to find the best linear transformation $\mathbf{a} \in \mathbb{R}^{p}$ and $\mathbf{b} \in \mathbb{R}^{q}$ that maximize the correlation: 
\begin{equation}
\operatorname{Cor}(\mathbf{U}, \mathbf{V})=\frac{\mathbf{a}^{T} \sum_{\mathbf{U} \mathbf{V}} \mathbf{b}}{\sqrt{\mathbf{a}^{T} \sum_{\mathbf{U} \mathbf{U}} \mathbf{a}} \cdot \sqrt{\mathbf{b}^{T} \sum_{\mathbf{V} \mathbf{V}} \mathbf{b}}},
\end{equation}
where
\begin{equation}
{\sum\mathop{}_{\mathbf{U} \mathbf{V}}}=\operatorname{Cov}(\mathbf{U},\mathbf{V})=\mathrm{E}[(\mathbf{U}-\mathrm{E}[\mathbf{U}])(\mathbf{V}-\mathrm{E}[\mathbf{V}])].
\end{equation}
The operator $\mathrm{E}$ denotes the expected value of its argument. This problem can be solved by SVD after basis change. 
This gives $\min(p,q)$ correlation coefficients sorted from the most correlated to the least correlated canonical directions. 
We consider the mean value $\overline{\operatorname{Cor}(\mathbf{U}, \mathbf{V})}$ of the correlation coefficients as an overall measure~\cite{svcca}.\par

\subsection{Facial Landmark Correlation}
 To focus on the variance of the face shape, we apply an important pre-processing step. We crop, center all the faces and then further normalize their sizes. We consider the 2D Cartesian coordinates as a two-dimensional variable.
Specifically, we calculate the absolute value of the correlation coefficients (ranged from -1 to 1) as we are interested in the magnitude of the correlation between two landmarks but not their directions. 
\par
To be clear, the canonical correlation between the $i$-th and the $j$ -th landmark in the annotation format can be found at the $i$-th row and the $j$-th column on the affinity matrix $\mathbf{A}$:
\begin{equation}
\mathbf{A}_{i,j}=\left| \, \operatorname{\overline{\operatorname{Cor}(\mathbf{L_{i}},\mathbf{L_{j}})}} \, \right|,
\end{equation}
where $\mathbf{L_{i}} , \mathbf{L_{j}} \in \mathbb{R}^{2}$ indicate the annotation of the $i$-th and the $j$-th landmark on the entire dataset. 
\par

The correlation affinity matrix on the 300W train subset~\cite{sagonas2013w300} is shown in Fig.~\ref{fig:teaser}. We draw several conclusions from these affinity matrices: \textbf{(\romannumeral 1)}  In general, the correlation among the landmarks belonging to the same facial component is more significant than the others. \textbf{(\romannumeral 2)} Some landmarks from the same component are weakly correlated (such as upper-lip and lower-lip). This is due to the shape variance of different facial expressions (such as mouth open/close). \textbf{(\romannumeral 3)} Certain landmarks from different facial components are strongly correlated, such as eyebrows and eyes, 
which is plausible. 

\section{Facial Landmark Model Interpretation}
\label{sec:CNN_interpretation}
In this section, we use the proposed landmark correlation analysis to gain some interesting insights on three important facial landmark detection models: cascaded random forest ERT~\cite{kazemi2014one}, cascaded CR-CNN~\cite{fan2016approaching} and HR-CNN~\cite{Bulat_2017_ICCV}. 
This section is divided into three parts:

\textbf{(\romannumeral 1)} The detailed settings of the three models that we interpret (Sect.~\ref{sec:4.1_model settings})

\textbf{(\romannumeral 2)} Some insights on the prediction of each model (Sect.~\ref{sec:4.2_characteristics})

\textbf{(\romannumeral 3)} The learning dynamics of the CR-CNN model (Sect.~\ref{sec:4.3_Network Learning Dynamics})

\begin{figure}[t]
\centering
\subfloat[Affinity Matrix Error (AME)]{\includegraphics[width=0.52\columnwidth]{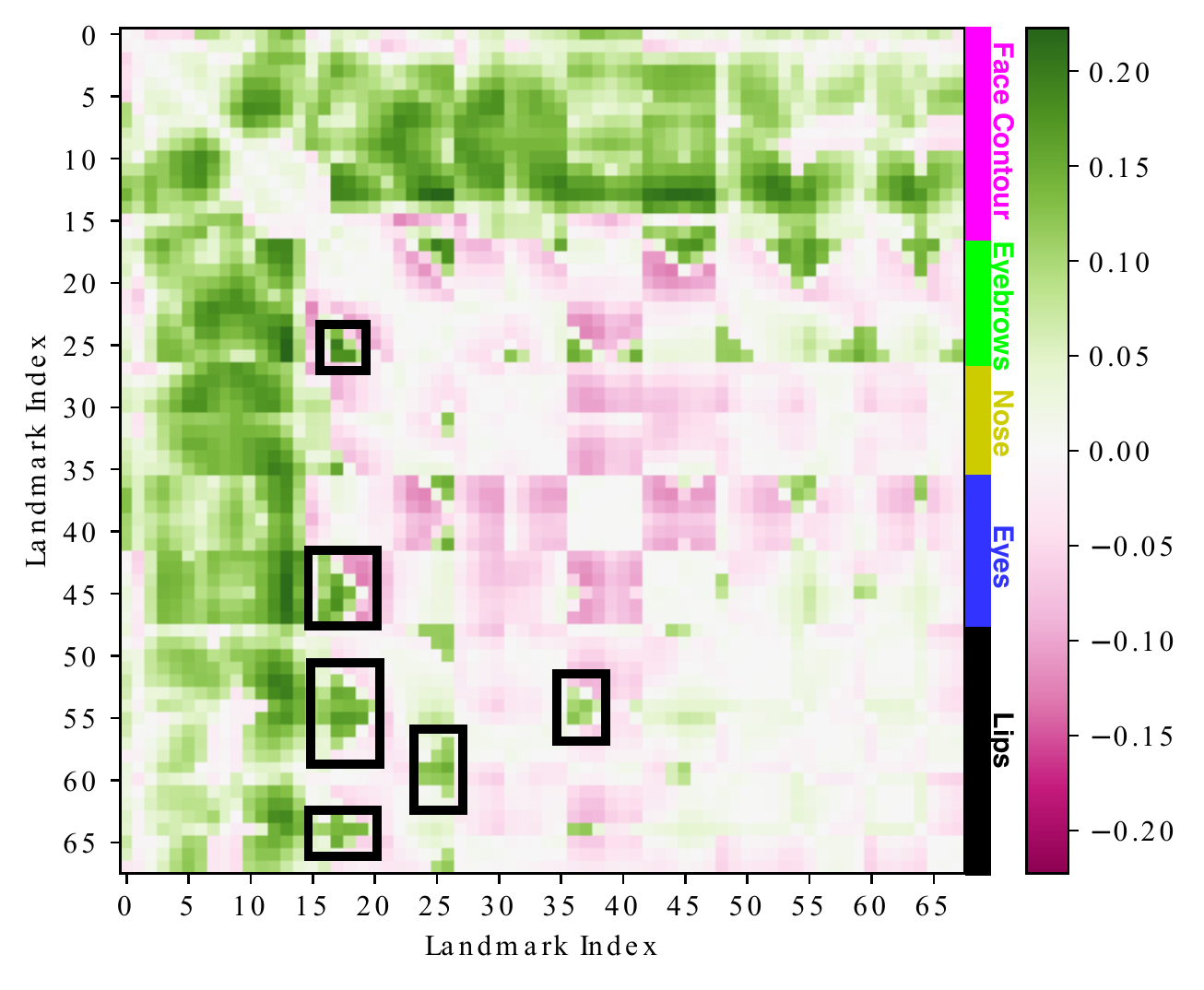}}
\subfloat[Examples of ERT prediction]{\includegraphics[width=0.39\columnwidth]{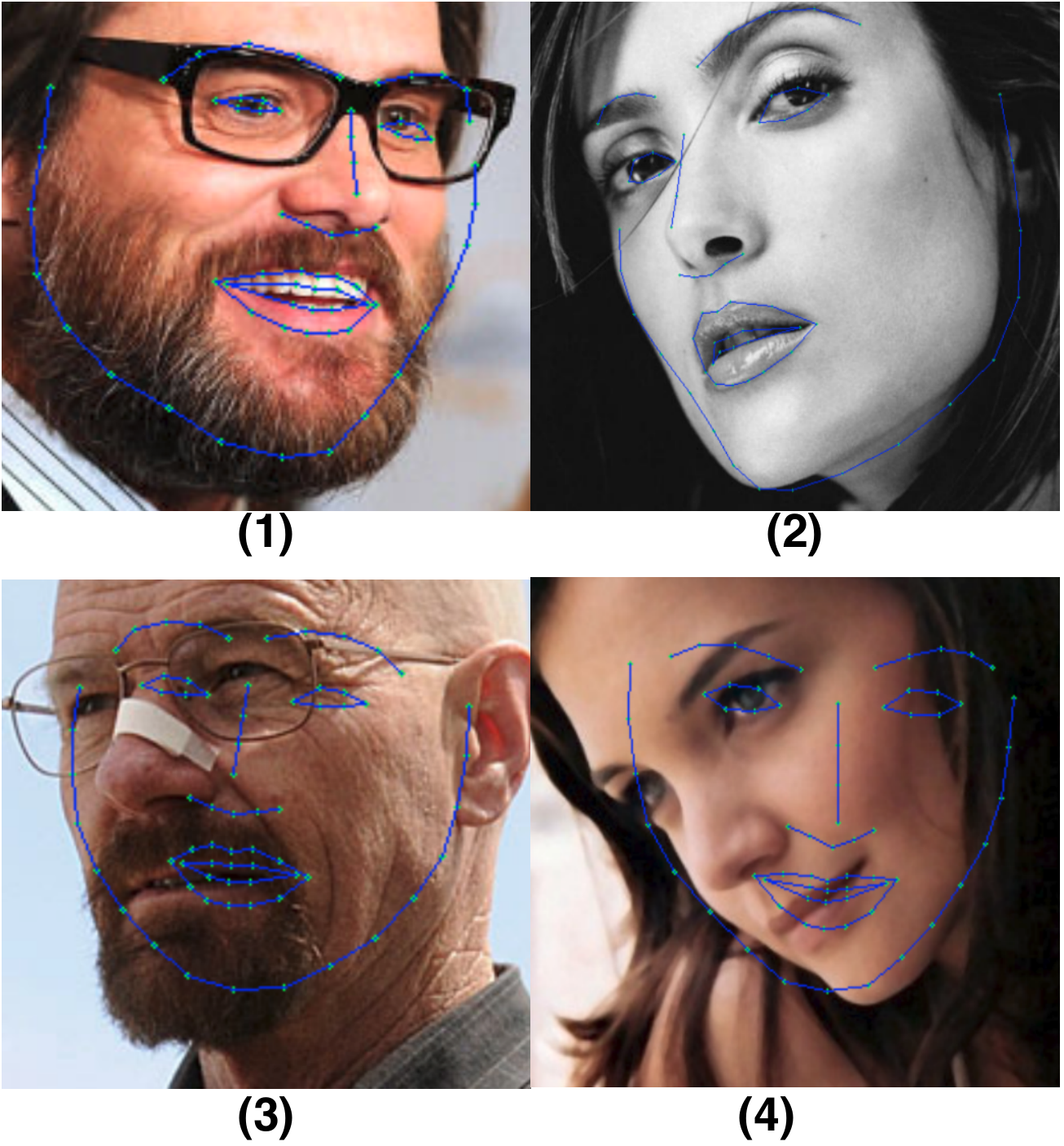}}
\caption{Demonstration of the final prediction from ERT. The rectangles in (a) highlight the correlation between some landmarks on the left \& right. For instance, the green color in the upper rectangle signifies the over correlation between the left eyebrow (N.17-N.21) and right eyebrow (N.22-N.26).}
\vspace{-5mm}
\label{fig:ert_final} 
\end{figure} 

\subsection{Model Settings}
\label{sec:4.1_model settings}
\par
\textbf{Cascaded Random Forest:} ERT~\cite{kazemi2014one} consists of 10 cascaded random forest regressors. Each regressor comprises 500 trees and the depth of the trees is 5. We use the implementation from~\cite{Xiao2019}. The initialized shape is the mean shape of the train subset. The NME of this model is 6.18\% on the validation set.
\par
\textbf{Cascaded CR-CNN:} We reproduced the model of Fan~\textit{et~al.}~\cite{fan2016approaching}. However, we added two additional stages to further boost its performance. Therefore, the overall structure has four stages. The main network in the first stage is ResNet18 and the sub-networks in the following stages consist of a single ResNet block and a single FC layer. The NME of this model is 3.66\%.
\par
\textbf{Stacked HR-CNN:} We used the official implementation of~\cite{Bulat_2017_ICCV}. The hourglass models are stacked in 4 stages. The NME of this model is 3.52\%.
\par
Unless specified, all of the models are trained and validated on 300W dataset.

\subsection{Characteristics of Each Model}
\label{sec:4.2_characteristics}
To compare the correlation difference between the prediction and the ground truth, we calculate the Affinity Matrix Error (AME), defined as:
\begin{equation}
AME = \mathbf{A}_{pred}-\mathbf{A}_{GT},
\end{equation} 
where $\mathbf{A}_{pred}$ is the CCA affinity matrix calculated on the prediction and $\mathbf{A}_{GT}$ is the CCA affinity matrix calculated on the ground truth.
In the following figures, green color indicates that the landmark correlation on the prediction is higher than the ground truth and red indicates lower correlation than the ground truth. 
\par

\textbf{Cascaded random forest ERT:} In Fig.~\ref{fig:ert_final} (a), we show the AME of the prediction from ERT. There are more green parts than red parts, and the shades of the green parts are higher than the red parts, which shows that the landmark correlation on the prediction of ERT is generally higher than the ground truth. 

We observe two important characteristics of ERT model:
\par
 \textbf{(\romannumeral 1)} The landmarks on the face contour are over-correlated to the other facial components, which means that the predicted face contour from ERT is too regularized. This observation can be confirmed by the visual examples (1) \& (2) in Fig.~\ref{fig:ert_final} (b).
\par
 \textbf{(\romannumeral 2)} Some landmarks on the right are over-correlated with other landmarks on the left (highlighted in the black rectangles in Fig.~\ref{fig:ert_final} (a)). For example, the correlation between the left eyebrow (landmark index N.17-N.21) and the right eyebrow (landmark index N.22-N.26) is significantly bigger than the ground truth. It statistically signifies that the prediction of the ERT does not present enough shape variance on the horizontal direction. We suspect that it is due to the failure confronting large head poses (see (3) \& (4) in Fig.~\ref{fig:ert_final} (b)). This characteristic can be confirmed in the literature~\cite{zhu2016unconstrained}. A widely acknowledged shortcoming of ERT model is its dependence on initialization, which brings poor robustness on the pose variance.
\par

\begin{figure}[t]
\centering
\subfloat[AME of Cascaded CR-CNN]{\includegraphics[width=0.5\columnwidth]{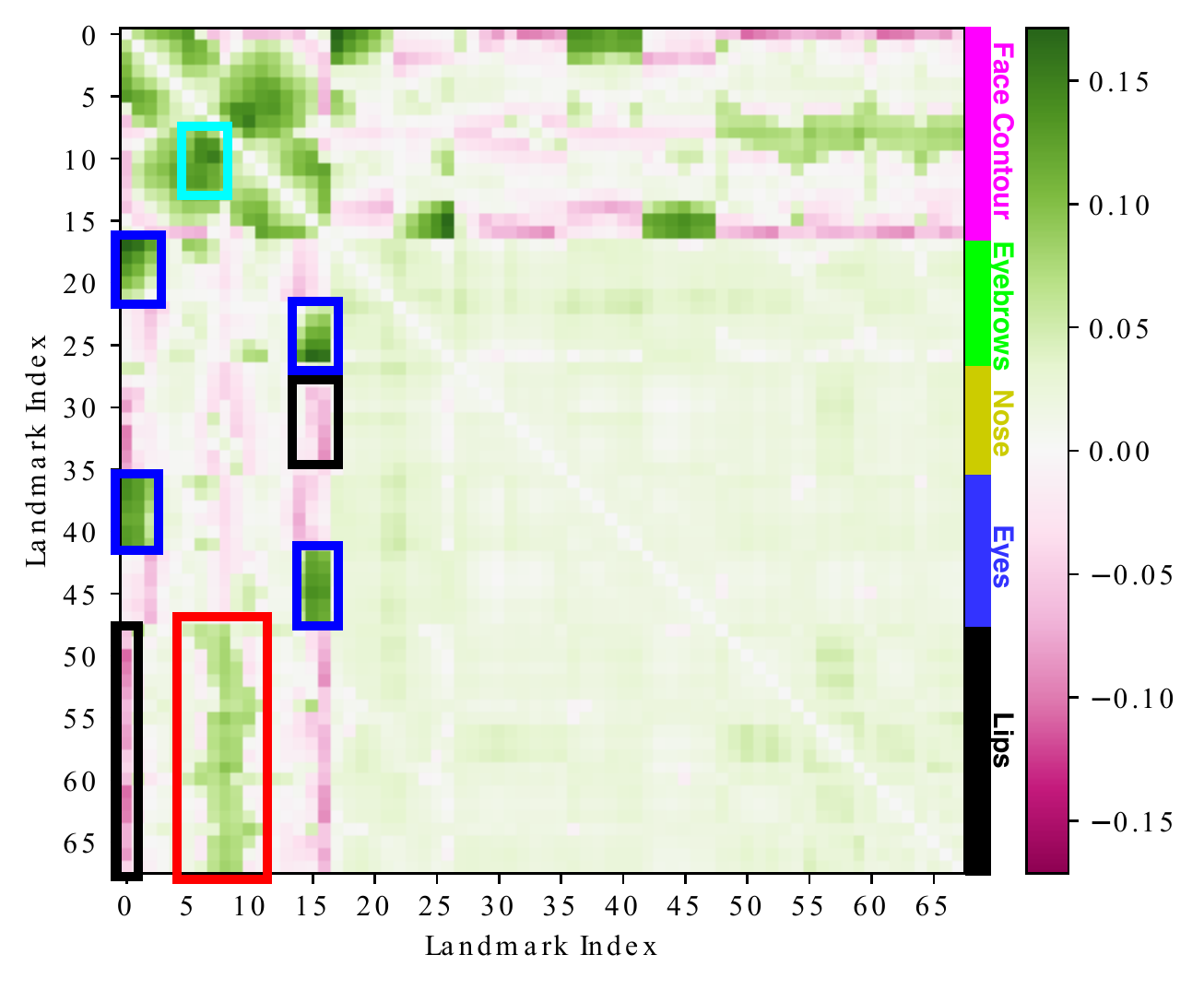}}
\subfloat[AME of Stacked HR-CNN]{\includegraphics[width=0.5\columnwidth]{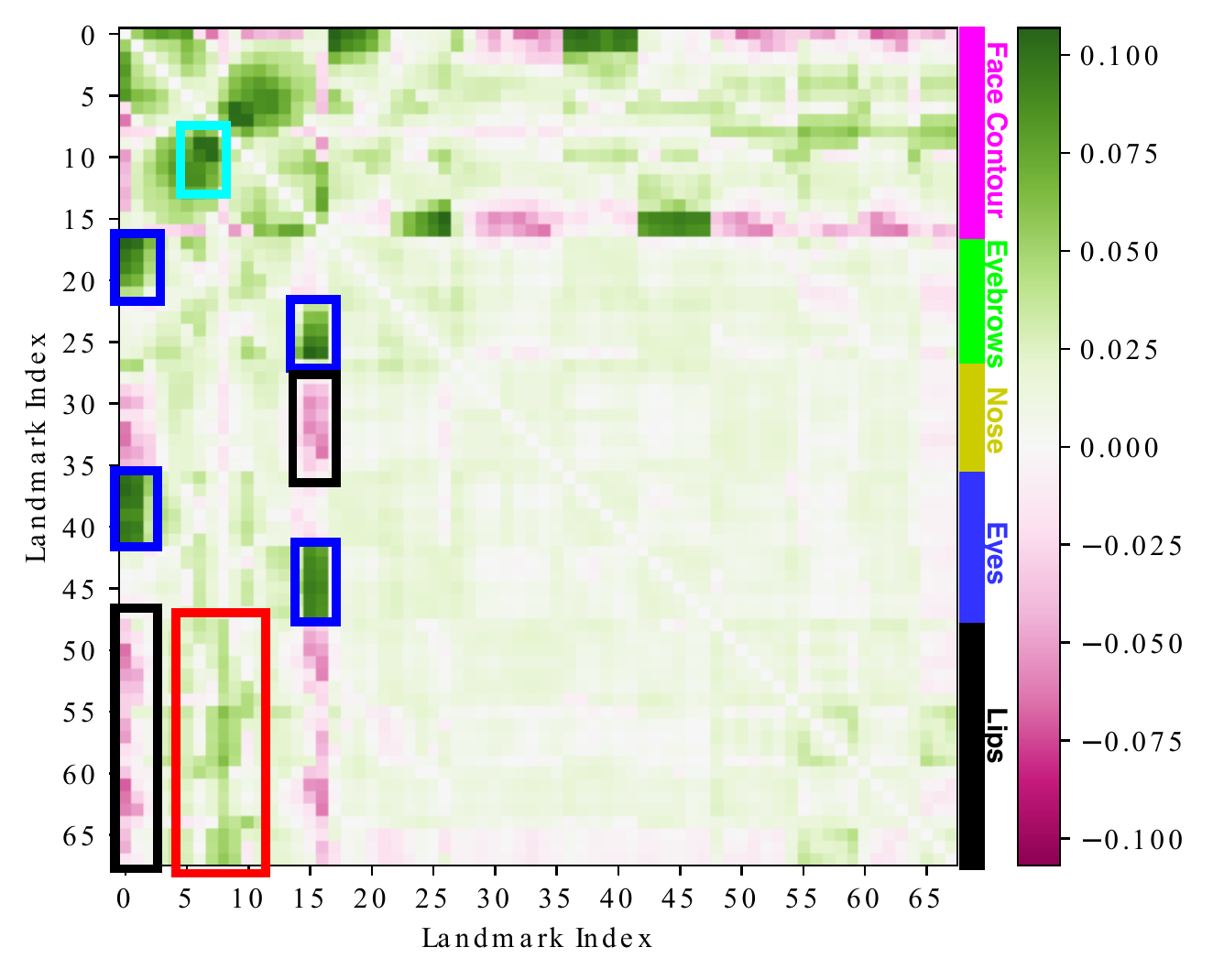}}
\caption{Demonstration of the final prediction from cascaded CR-CNN and stacked HR-CNN. For example, the lower right blue rectangle highlights that the upper right face contour (landmark N.14 - N.16) and the right eye (N.42 - N.47) are over-correlated, on both CNNs. Similarly, the red rectangle highlights that the lower face contour (N.5 - N.11) and the lips (N.48 - N.67) are over correlated.}
\label{fig:CNN_final} 
\end{figure}

\textbf{Cascaded CR-CNN \& stacked HR-CNN:} The AMEs of CR-CNN and HR-CNN are shown in Fig.~\ref{fig:CNN_final}. In general, the prediction of stacked HR-CNN presents a lower correlation error compared to cascaded CR-CNN (see the scale of colorbar on the right). 
The correlation error mainly exists on the face contour.

We note one similarity and two differences between two CNN models:

\textbf{Similarity: Both of the CNNs tend to correlate adjacent landmarks.} For example, in Fig.~\ref{fig:CNN_final}, the correlation between the eyebrows/eyes and the side face contour (blue rectangles), the correlation between the lips and the bottom face contour (red rectangles) and the correlation among the landmarks on the bottom face contour (cyan rectangles) are significantly higher than the ground truth correlation. 
Some landmarks that are already over-correlated to their adjacent landmarks, show inferior correlation with the distant landmarks (the correlation between upper-left face contour and lips, black rectangle). 
This is probably due to the convolution operation used in the CNN, which excessively exploits local semantic information.
\par

\begin{figure}[t]
\centering
\subfloat[AME of Single-stage CR-CNN]{\includegraphics[width=0.495\columnwidth]{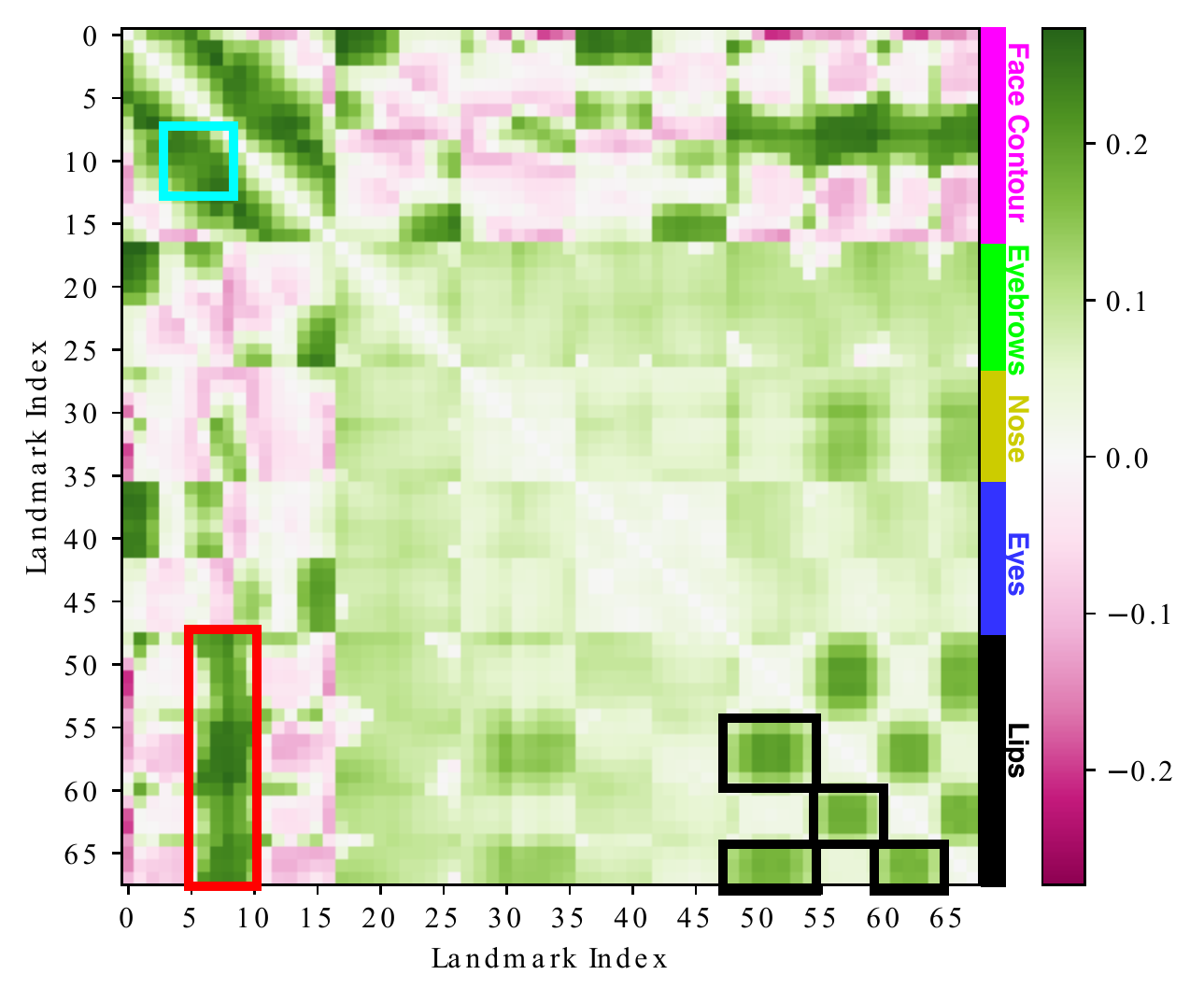}}
\subfloat[AME of Single-stage HR-CNN]{\includegraphics[width=0.505\columnwidth]{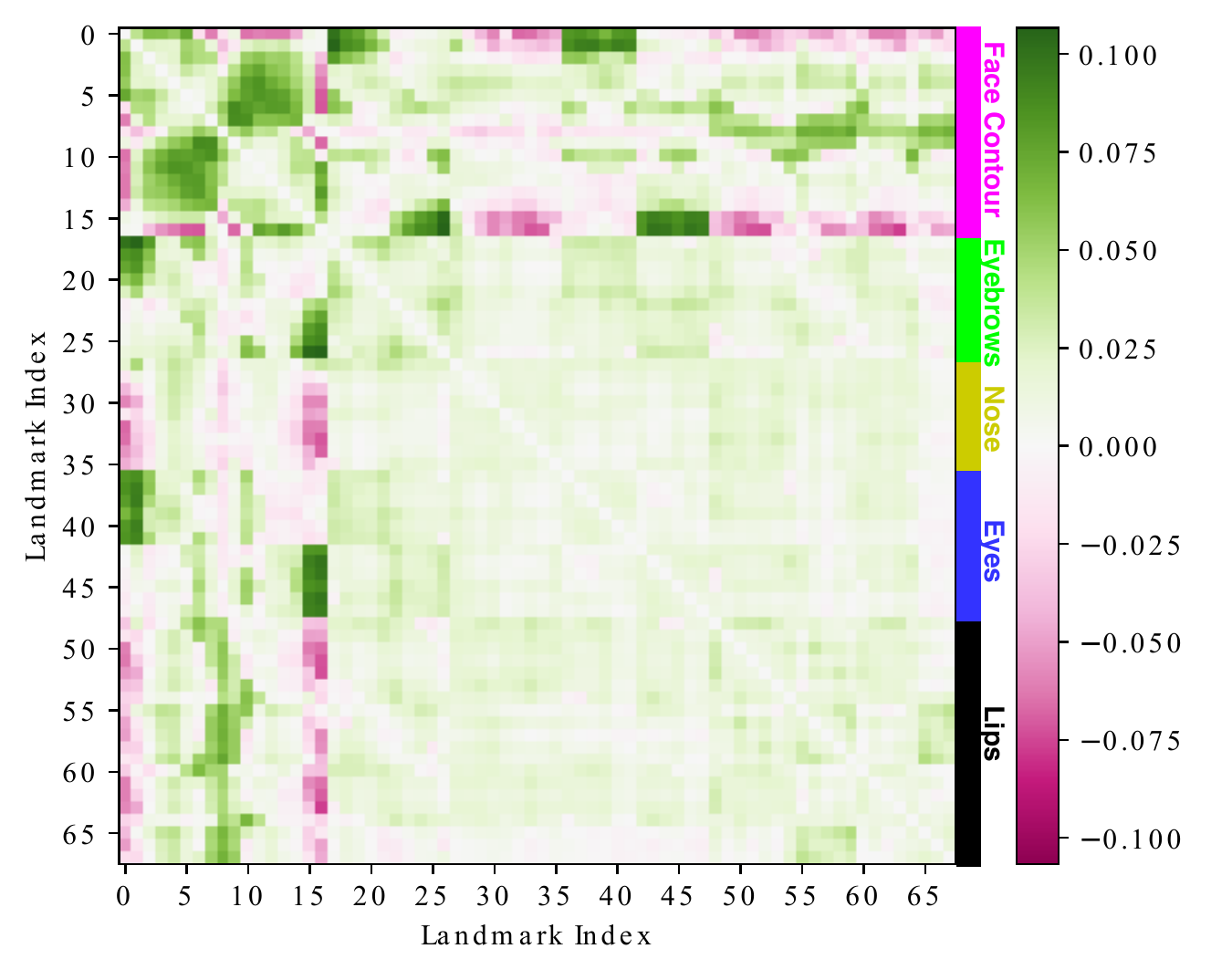}}
\caption{Demonstration of the final prediction from single-stage cascaded CR-CNN and single-stage HR-CNN. In (a), we find that the correlation between the upper lip and the lower lip (black rectangles), the lips and lower face contour (red rectangle) and the correlation among the adjacent landmarks on the face contour (cyan rectangle) are significantly higher than those in Fig.~\ref{fig:CNN_final} (a).}
\vspace{-4mm}
\label{fig:singlestage} 
\end{figure} 

\par

\textbf{Difference 1: The coarse-to-fine strategies in HR-CNN (stacking) and CR-CNN (cascading) are different.} In Fig.~\ref{fig:singlestage}, we demonstrate the AME of single-stage HR-CNN and CR-CNN. By comparing Fig.~\ref{fig:singlestage} (a) and Fig.~\ref{fig:CNN_final} (a), we observe that a single-stage CR-CNN suffered from more severe over-correlation problem compared to the cascaded CR-CNN (see the black, red and cyan rectangles in Fig.~\ref{fig:singlestage} (a)). 
Therefore, the coarse prediction of the first stage in cascaded CR-CNN is indeed over-regularized. 
The following cascaded stages learn the further shape variance, which de-correlates the output shape from the first stage. We note that in CR-CNNs, the output is linearly connected with the previous fully-connected layer. 
This may explain why there are always excessive correlations present on the output of CR-CNN. 
Comparing Fig.~\ref{fig:singlestage} (b) and Fig.~\ref{fig:CNN_final} (b), we find that the correlation error of the prediction from single-stage HR-CNN is similar to the stacked HR-CNN. Therefore, we think that the role of stacking in HR-CNN is different from the role of cascading in CR-CNN. Further study on this issue is presented in the supplementary material.
\par

\begin{figure}[t]
\centering
\subfloat[AME of Cascaded CR-CNN]{\includegraphics[width=0.495\columnwidth]{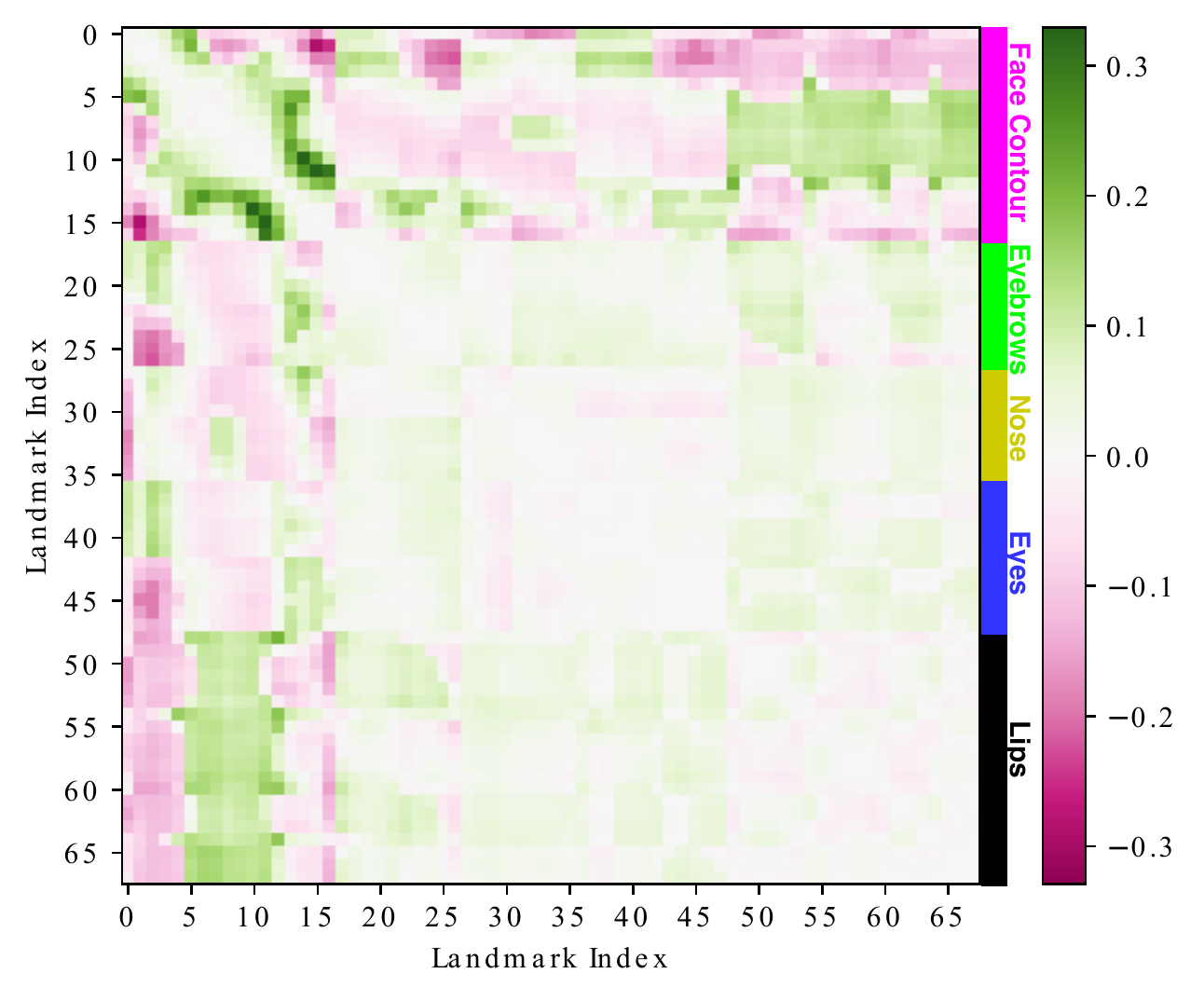}}
\subfloat[AME of Stacked HR-CNN]{\includegraphics[width=0.505\columnwidth]{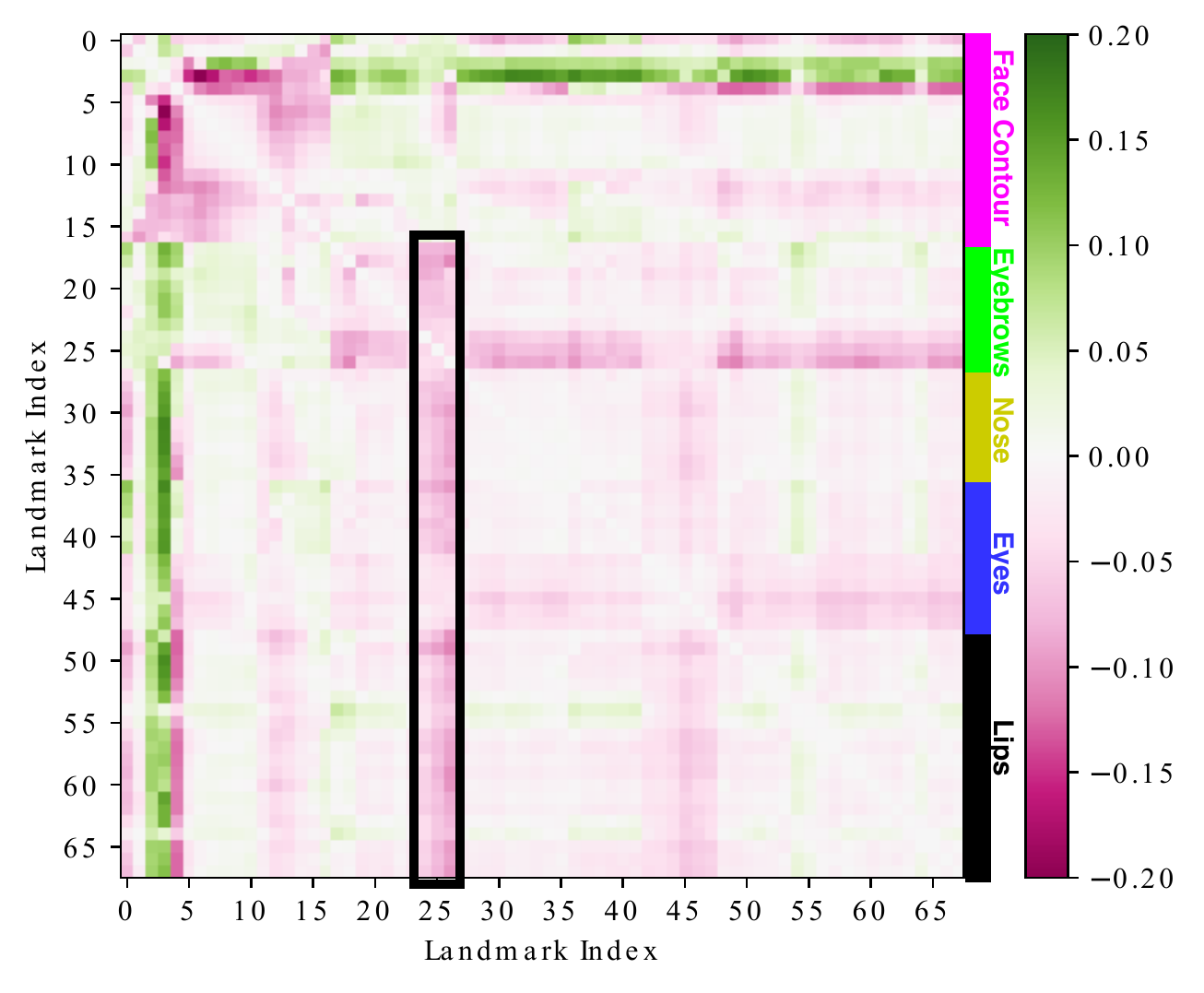}}
\caption{The AME of cascaded CR-CNN and stacked HR-CNN on 300VW Scenario 3. The rectangle in (b) highlights the inferior landmark correlation than the ground truth between right eyebrow (N.22 - N.26) and other facial components.}
\label{fig:VWS3} 
\end{figure} 

\textbf{Difference 2: HR-CNN is more likely to violate the landmark correlation than CR-CNN under challenging conditions.} In Fig.~\ref{fig:VWS3}, we demonstrate the AME of two CNN models on a challenging video dataset 300VW S3~\cite{shen2015first}, which involves noisy conditions such as occlusions, motion blurs, complex lighting conditions, etc.. For stacked HR-CNN, we find that the correlation between the inner facial components is weaker than the ground truth, especially on the right eyebrow (black rectangle in Fig.~\ref{fig:VWS3} (a)). 
This is consistent with the weakness of HR-CNN that we mentioned in Fig.~\ref{fig:weakness} (b). 
However, if we compare Fig.~\ref{fig:VWS3} (a) and Fig.~\ref{fig:VWS3} (b), we find that CR-CNN is still robust under these challenging conditions, especially on the landmarks of inner facial components.
\par

\subsection{CR-CNN Learning Dynamics} 
\label{sec:4.3_Network Learning Dynamics}

In this section, we study how the CR-CNN progressively learns from the beginning. To this end, we plot the evolution of the CR-CNN output correlations during training. We do not analyze the learning dynamic of HR-CNN due to the operation of taking the maximum value on the final heatmap. We think that it requires a 2D CCA method in the future to analyze the output heatmap directly.
\par

We trained a ResNet-18 on the 300W dataset. Both the convolutional layers and the FC layers are initialized from a normal distribution~\cite{he2015delving}. The model is trained for 350 epochs with the learning rate decayed by 0.3 for each 70 epochs. We observe the following phases during the first 70 epochs:
\par
\begin{figure}[t]
\centering
\subfloat[Epoch 0: 170.1\%]{\includegraphics[width=0.25\textwidth]{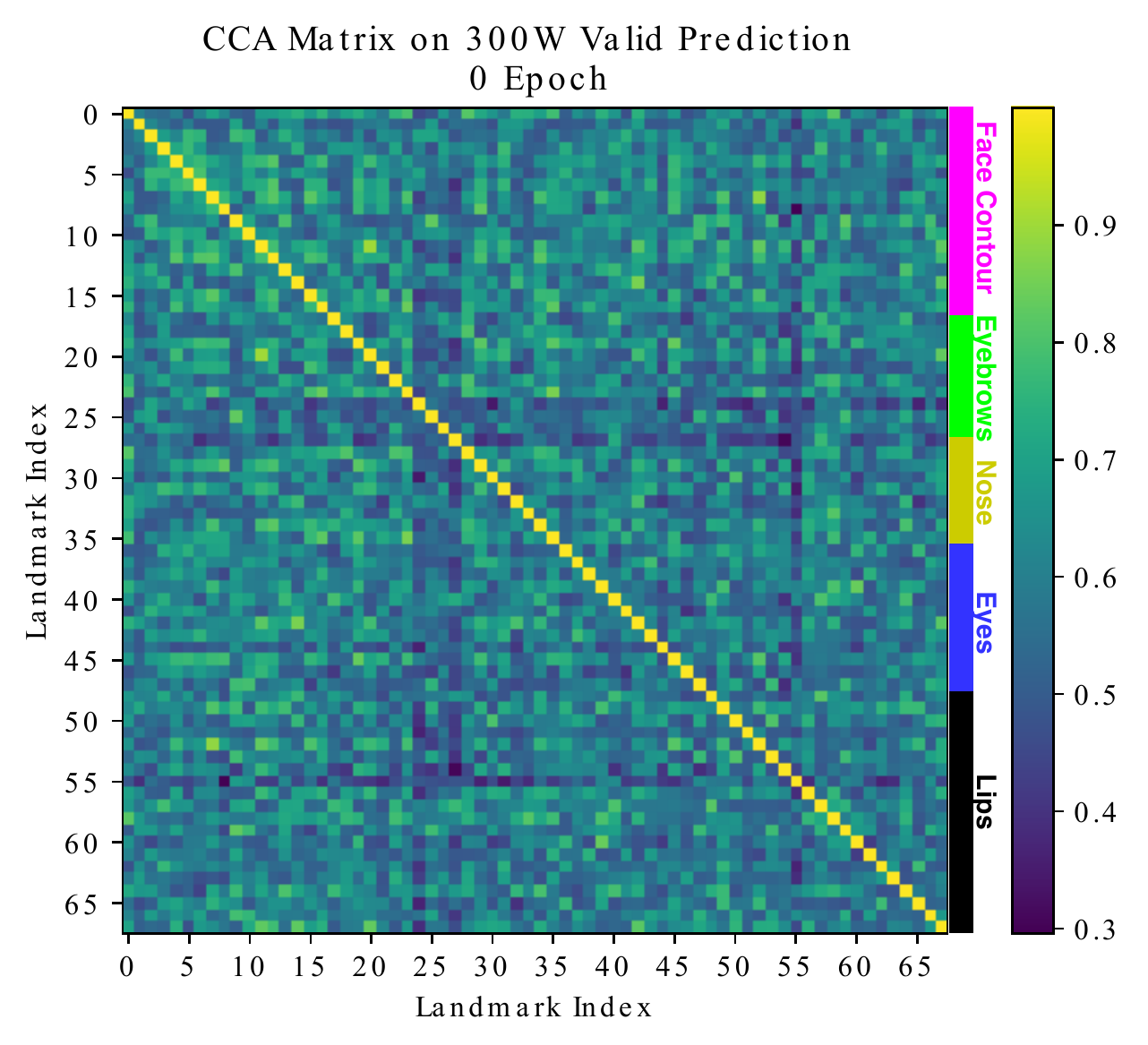}}
\subfloat[Epoch 12: 15.5\%]{\includegraphics[width=0.25\textwidth]{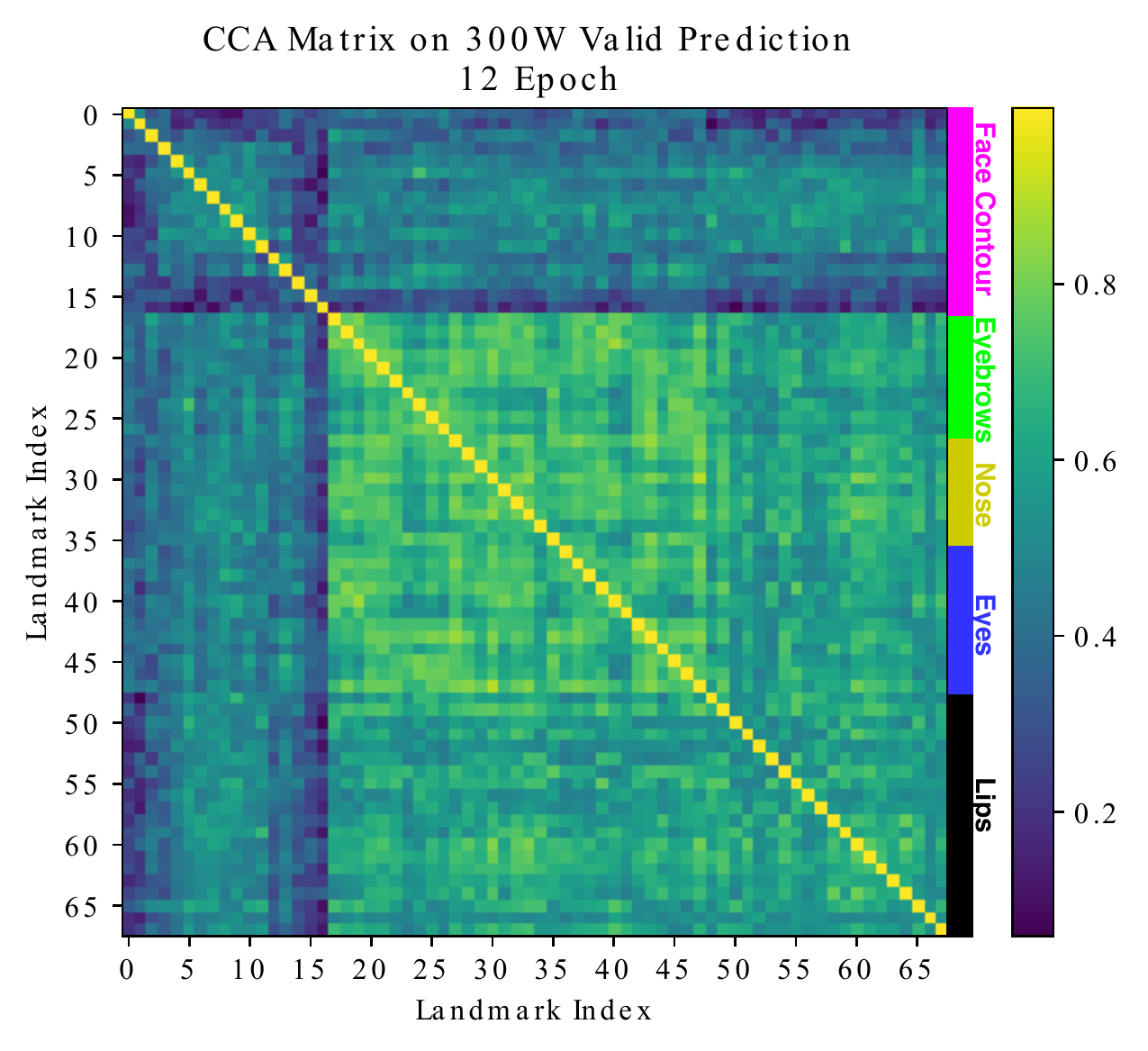}}
\subfloat[Epoch 20: 10.0\%]{\includegraphics[width=0.25\textwidth]{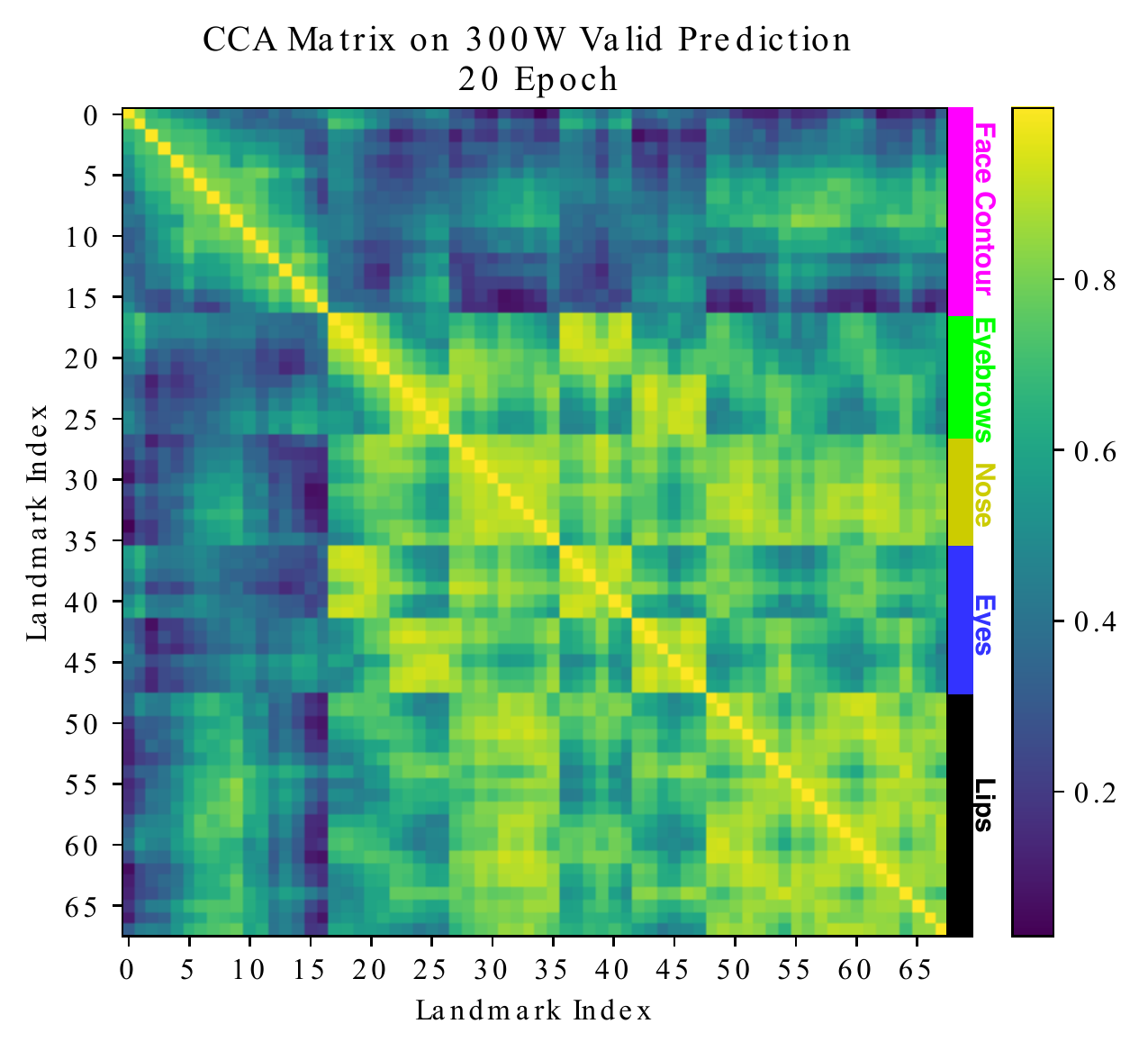}}
\subfloat[Epoch 70: 6.6\%]{\includegraphics[width=0.25\textwidth]{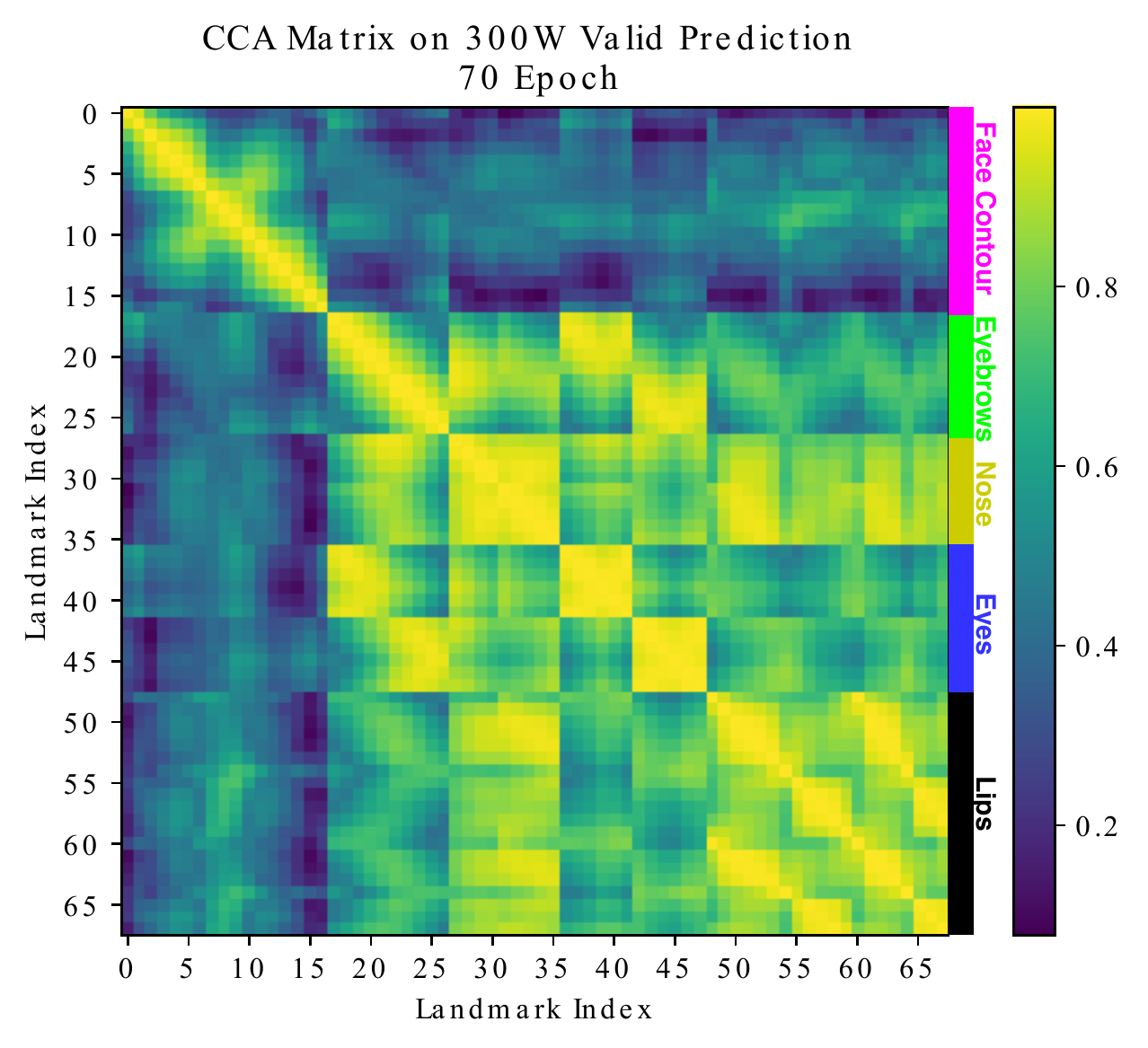}}
\caption{CCA affinity matrices on the prediction of CR-CNN in different training epochs. The percentage shown in each figure caption refers to NME.}
\vspace{-4.5mm}
\label{fig:lrdy} 
\end{figure}

\textbf{Phase 1 Group Inner Facial Components:} More rigid parts learn first. The first thing that CNN starts to learn is to group the inner facial components. We can observe in Fig.~\ref{fig:lrdy} (b) that CNN firstly learns a relatively strong correlation among the landmarks on the inner facial components (eyebrows, eyes, noses) and separate them from the other landmarks on face contours. 
\par
\textbf{Phase 2 Recognize Each Facial Component:} Next, the CNN starts to gradually identify the facial components (eyebrows, nose, mouths, etc.). In this phase, the correlation among the landmarks which belong to the same facial component grows stronger (see Fig.~\ref{fig:lrdy} (c)). The CNN recognizes the eyes, nose and lips almost simultaneously.
\par
\textbf{Phase 3 Refine the Prediction:} The CNN learns to refine the prediction in two aspects: \textbf{(\romannumeral 1)} enhance the correlation inside each facial component, especially the neighbouring landmarks; \textbf{(\romannumeral 2)} reduce some excessive correlations (e.g. the correlation between lower face contour and lips, see Fig.~\ref{fig:lrdy} (c) \& (d)).\par

\begin{figure}[t]
\centering
\subfloat[Epoch 71-140: 5.7\%]{\includegraphics[width=0.4\columnwidth]{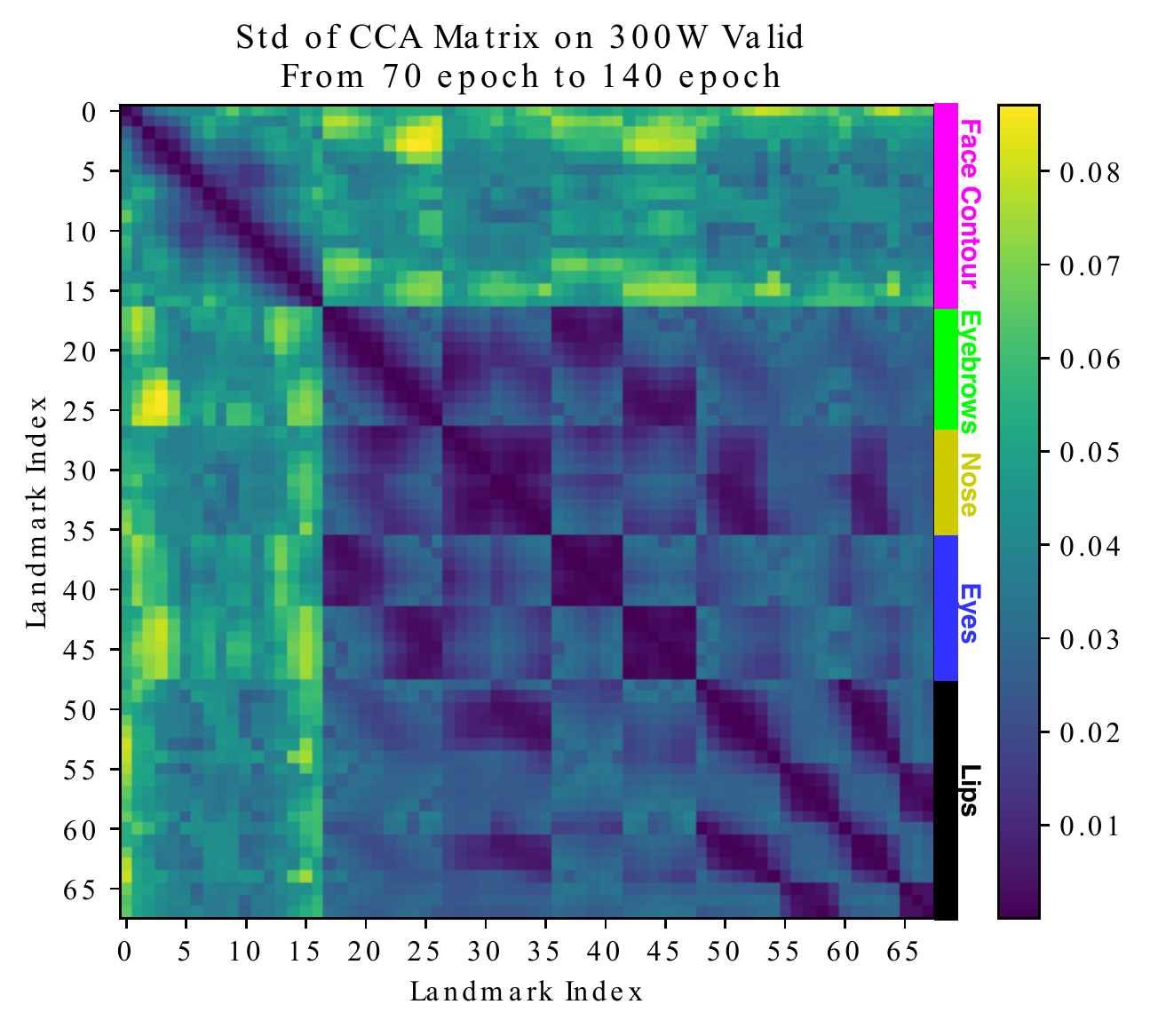}}
\subfloat[Epoch 141-280: 4.8\%]{\includegraphics[width=0.4\columnwidth]{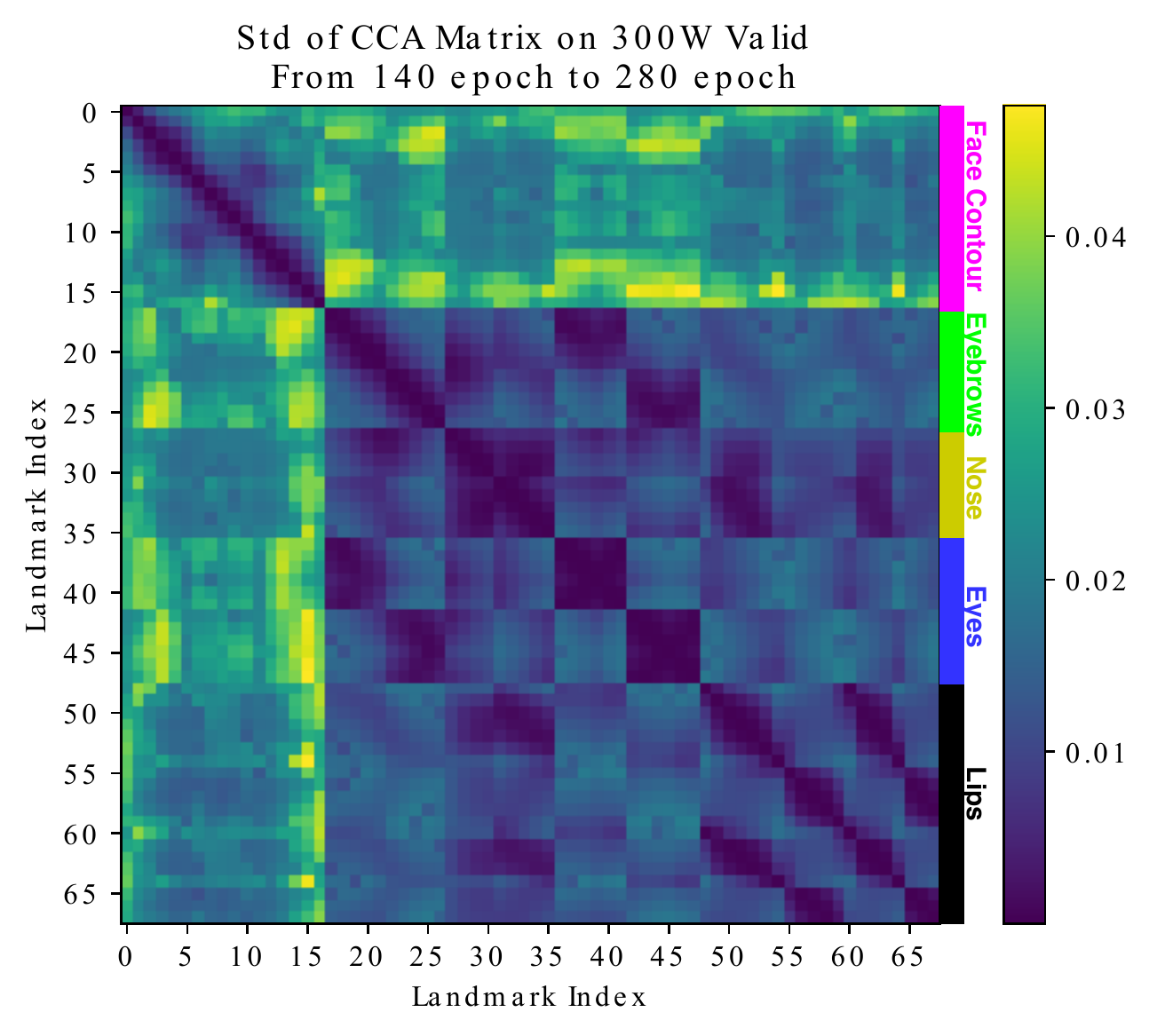}}
\caption{Standard deviation (Std) of CCA affinity matrices on the prediction of CR-CNN in different training epochs. The percentage shown in each caption refers to NME at 140th/280th epoch respectively.}
\label{fig:std} 
\end{figure}

The evolution of the affinity matrices after 70 epochs is difficult to visualize as the evolution of the correlation value is small. Therefore, we calculate the standard deviation (Std) of the affinity matrices in different stages (see Fig.~\ref{fig:std}). We observe that after 70 epoch, the variation of the correlation value related to the landmarks on the face contour is significantly higher than the others, indicating that the model struggles to refine the face contour.\par

\section{Few-short learning}

\label{sec:Few-shot Learning}

\textbf{Motivation:} As the size of datasets grows larger and the landmark format becomes denser~\cite{wu2018look}, it is time-consuming to densely annotate each landmark on all of the images. Few-shot learning has attracted increasing attention in the community. Due to the presence of strong landmark correlation in the dense format, we believe that it is not cost-effective to annotate every landmark, especially when the budget for manual annotation is limited. \par

Most existing landmark transfer methods use existing sparse formats~\cite{belhumeur2013LFPW,zhang2016MAFL,van2012LFW,burgos2013COFW,Koestinger2011AFLW} to infer dense formats. 
However, it is not ensured that the existing sparse formats are optimal for the underlying data. 
Therefore, we propose to \textit{search for} a sparse format, by selecting a set of landmarks which are most correlated to the rest of them. 
We assume that a landmark can be easily transferred from another landmark that is highly correlated.
Our method shows two advantages: 
\textbf{(\romannumeral 1)} The form of our sparse format is purely data-driven.
\textbf{(\romannumeral 2)} The number of landmarks in the sparse format can be arbitrarily chosen depending on the minimum correlation required between the selected landmarks and the rest.\par

\begin{figure}[t]
\begin{center}
\includegraphics[width=\textwidth]{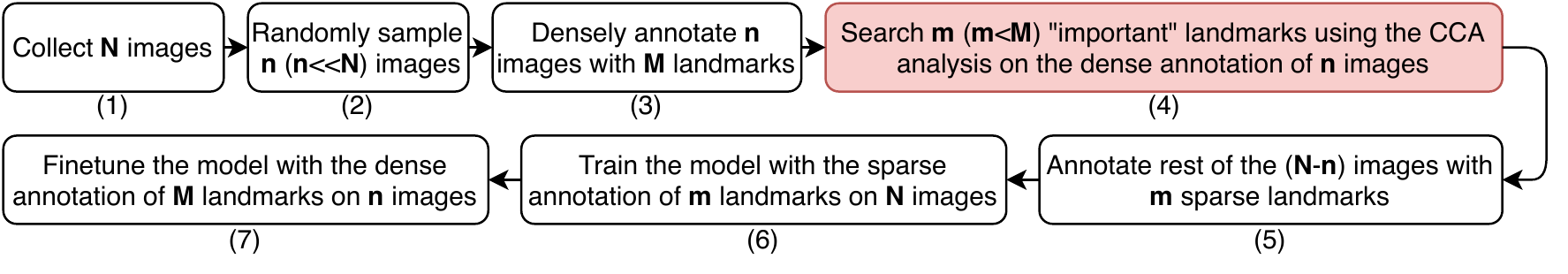}
\end{center}
   \caption{The workflow of our few-shot learning method. \textbf{M}: total number of the landmarks in the dense format \textbf{N}: total number of the images collected. \textbf{m}: number of the landmarks in the sparse format, can be considered as annotation budget. The value of \textbf{m} and \textbf{n} can be arbitrarily chosen. We save the time to annotate \textbf{M}-\textbf{m} landmarks on \textbf{N}-\textbf{n} images.}
\label{fig:fewshot_workflow}
\vspace{-0.8mm}
\end{figure}

\textbf{Workflow:} We believe that the landmark correlation can be analyzed on a small batch. Therefore, only a portion of images need to be densely annotated for searching the sparse format. The workflow of our few-shot regression method is shown in Fig.~\ref{fig:fewshot_workflow}. 
The search of sparse annotation format is done in step (4), which will be discussed in the next subsection.

Our method is also useful for extending a dense landmark format, without starting from the very beginning. It is practical for individuals or small teams who want to develop their own dense format for a specific use. For example, we want to detect the landmarks on the wing of the nose or the landmarks around the forehead, which are not annotated in the 300W format.  We can first analyze on a small batch how these new landmarks are correlated to the landmarks already annotated, then find out an efficient strategy for manual annotation.\par

\textbf{Problem Formulation:} We formulate Fig.~\ref{fig:fewshot_workflow} step (4) as a Maximin problem: 
Find a set of ``important'' landmarks indexed by $\mathit{m}$, which maximize the minimum correlation $\mathbf{c}$ with rest of the landmarks indexed by  $\mathcal{M}-\mathit{m}$: 

\begin{equation}
\mathit{m} = \argmax_{\mathit{m} \subset \mathcal{M}}(\mathbf{c}_{\mathit{m}}),
\label{fom:min_cor}
\end{equation}

\begin{equation}
\mathbf{c}_{\mathit{m}} = \min\nolimits_{\mathit{j} \in (\mathcal{M}-m)}(\max\nolimits_{\mathit{i} \in \mathit{m}}(\mathbf{A}_{i,j})).
\label{fom:hat_c}
\end{equation}

$\mathcal{M}$ denotes the complete set of landmark index in the dense format. $\mathbf{A}_{\mathit{i}, \mathit{j}}$ denotes the $\mathit{i}$-th row and $\mathit{j}$-th column of the correlation affinity matrix $\mathbf{A}$ analyzed on $\textbf{n}$ images. 
The maximized minimum correlation $\mathbf{\hat{c}}$ can be obtained by $\mathbf{\hat{c}}=\max_{\mathit{m} \subset \mathcal{M}}(\mathbf{c}_{\mathit{m}})$. 
\par

\textbf{Solving the Maximin Optimization:} This problem resembles K-center facility problem~\cite{hochbaum1985best}. A classical K-center problem is described as: Given a city with \textbf{M} locations, find the best k locations to build k facilities, so that the farthest distance from location to its nearest facility has to be as small as possible. 
\par
In our problem, the locations in the city can be considered as all the landmarks in the dense annotation format $\mathcal{M}$. The k facilities can be considered as the landmarks selected in our sparse format $\mathit{m}$. The distance between the landmark $i$ and $j$ can be considered as $1-\mathbf{A}_{i,j}$. In fact, a high correlation between two landmarks signifies that the distance between two landmarks is small.
\par
K-center problem is NP-hard. Fortunately, this problem can be efficiently solved by mixed-integer programming using Gurobi~\cite{gurobi}, a powerful mathematical optimization solver. 
We present the canonical form of this problem:
\par
Minimize $z$, with subject to:
\begin{equation}
\begin{array}{ll}{\sum_{j} x_{i j}=1} & {\forall i} \\ {\sum_{j} y_{j}=\textbf{m}} & {} \\ {x_{i j} \leq y_{j}} & {\forall i, j} \\ {(1-\mathbf{A}_{i,j})x_{i j} \leq z} & {\forall i, j} \\ {x_{i j} \in\{0,1\}} & {\forall i, j} \\ {y_{j} \in\{0,1\}} & {\forall j}.\end{array}
\end{equation}
$x_{i j}=1$ indicates that landmark $i$ is inferred from the position of landmark $j$. $y_{j}=1$ indicates that the landmark $j$ is selected in the sparse format. $\sum_{j} x_{i j}=1$ ensures that all the landmarks are inferred from another landmark. $\sum_{j} y_{j}=\textbf{m}$ ensures that there are \textbf{m} landmarks selected in the sparse format. $x_{i j} \leq y_{j}$ ensures that landmark $i$ can be inferred from landmark $j$ only when landmark $j$ is selected in the sparse format. Finally, the maximized minimum correlation can be obtained by $\mathbf{\hat{c}}=1-z$. This optimization can be finished in just several seconds on a normal computer.
\par

\begin{figure}[t]
  \centering
  \includegraphics[width=0.9\textwidth]{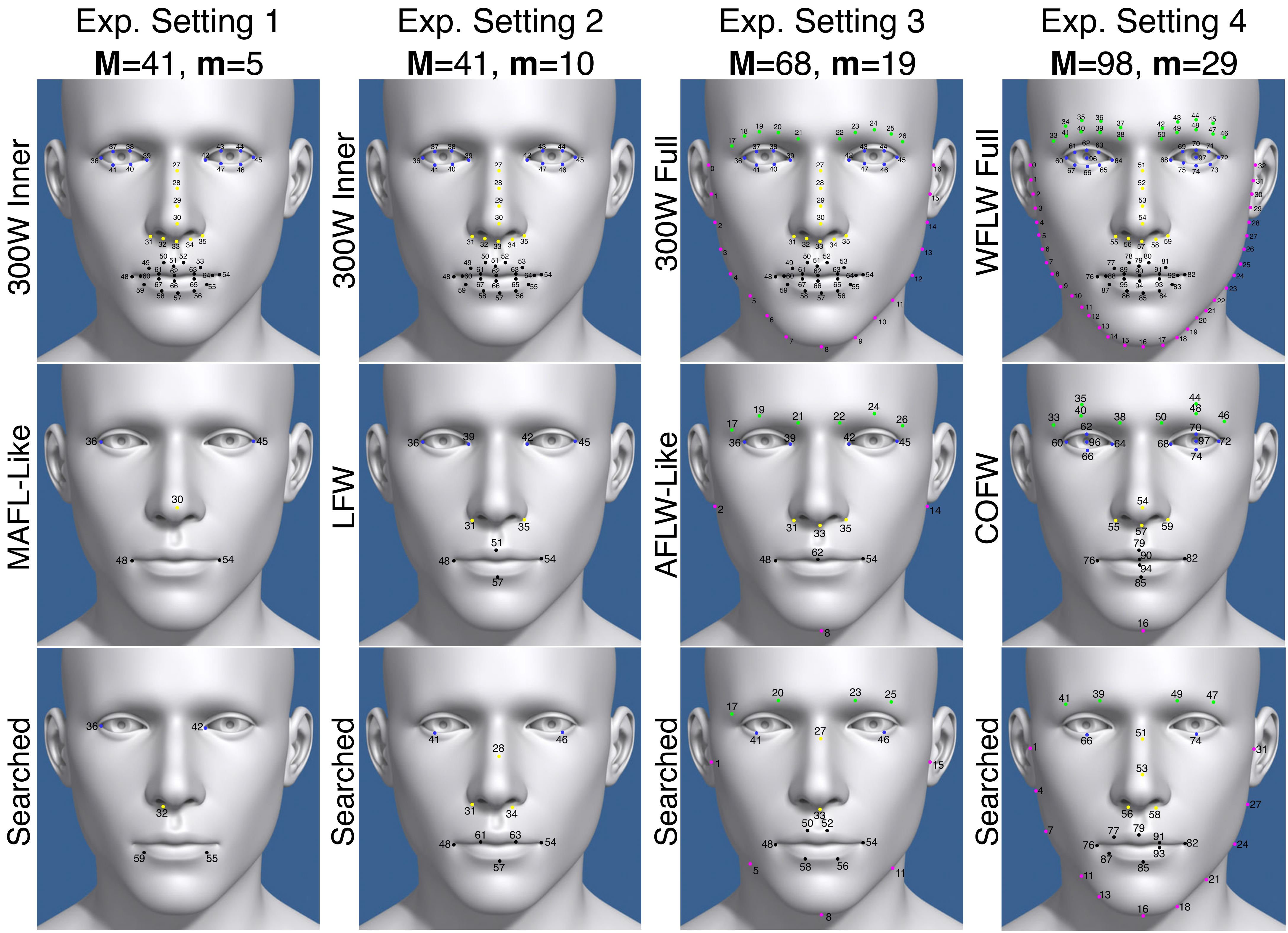}
  \caption{Demonstration of experimental settings. The first row shows the dense format with \textbf{M} landmarks. The second row shows an existing sparse format with \textbf{m} landmarks. The third row shows one of the sparse formats searched by our method with \textbf{m} landmarks. Each column represents an experimental setting that we perform in Tab.~\ref{tab:few_shot_comp}. The number of dense landmarks \textbf{M} and the number of sparse landmarks \textbf{m} are indicated on the top of each column.}
  \label{fig:searched_index}
  \vspace{-3mm}
\end{figure}

\textbf{Experimental Settings:} In the experiments, we want to prove that the searched sparse formats are more efficient than existing sparse formats. We propose four settings, demonstrated by column in Fig.~\ref{fig:searched_index}. In each setting, we search identical number of sparse landmarks (annotation budget) \textbf{m} as existing formats.
(\romannumeral 1) Search 5 sparse landmarks out of 41 dense landmarks on 300W Inner, compared with MAFL~\cite{zhang2016MAFL} format (the 1st column).
(\romannumeral 2) Search 10 sparse landmarks out of 41 dense landmarks on 300W Inner, compared with LFW~\cite{van2012LFW} format (the 2nd column).
(\romannumeral 3) Search 19 sparse landmarks out of 68 dense landmarks on 300W Full, compared with AFLW~\cite{Koestinger2011AFLW} format (the 3rd column).
(\romannumeral 4) Search 29 sparse landmarks out of 98 dense landmarks on WFLW Full, compared with COFW~\cite{burgos2013COFW} format (the 4th column).
We also adopted different image sampling ratios \textbf{n}/\textbf{N} from 5\% to 25\% (concerned in Fig.~\ref{fig:fewshot_workflow} step(2)). For each setting, we run our searching method for three times and present the average value.

On the 3rd row of Fig.~\ref{fig:searched_index}, we demonstrate one of the sparse formats searched by our method. Note that the searched formats can be different each time depending on the data (\textbf{n} images) sampled from the entire dataset (\textbf{N} images). Visually, we find that our method is able to distribute the annotation budget (\textbf{m} landmarks to annotate) more evenly on each part.

\par

\textbf{Results:} In Tab.~\ref{tab:few_shot_comp}, we present the performance comparison on this task. Our sparse format achieves comparable performance compared to LFW format and slightly better results compared to MAFL format on 300W Inner. When the landmarks on the face contour are included in the learning (on 300W Full \& WFLW Full), our format demonstrates more significant improvement compared to AFLW format and COFW format. We also noticed that our searched format is more advantageous with lower sampling ratio \textbf{n}/\textbf{N} (in other words with fewer densely annotated images). NME difference between our format and pre-defined format is much larger when trained with ratio of 5\% and 10\%.\par

\begin{table}[t]
\centering
\begin{tabular}{l|ll|ll|ll|ll}
\toprule
        & \multicolumn{2}{c|}{Exp. Setting 1} & \multicolumn{2}{c|}{Exp. Setting 2} & \multicolumn{2}{c|}{Exp. Setting 3} & \multicolumn{2}{c}{Exp. Setting 4}\\
        & \multicolumn{2}{c|}{\textbf{M}=41, \textbf{m}=5} & \multicolumn{2}{c|}{\textbf{M}=41, \textbf{m}=10} & \multicolumn{2}{c|}{\textbf{M}=68, \textbf{m}=19} & \multicolumn{2}{c}{\textbf{M}=98, \textbf{m}=29} \\\hline
 Ratio (\textbf{n}/\textbf{N}) & MAFL~\cite{zhang2016MAFL}     & Ours     & LFW~\cite{van2012LFW}         & Ours         & AFLW~\cite{Koestinger2011AFLW}    & Ours        & COFW~\cite{burgos2013COFW}         & Ours        \\ \hline
5\%  &      4.24     &    \textbf{4.17}       &    \textbf{3.80}    &     3.83         & 5.32         & \textbf{5.17}   &    7.27    &     \textbf{6.99}      \\ \hline
10\%  &      3.92     &    \textbf{3.86}       &    \textbf{3.59}    &    3.61         & 5.08         & \textbf{4.94}   &    7.03   &     \textbf{6.70}      \\ \hline
25\%  &      3.74     &    \textbf{3.66}       &    3.43    &     3.43         & 4.85        & \textbf{4.84}   &    6.62    &     \textbf{6.42}      \\ \bottomrule
\end{tabular}
\caption{NME(\%) performance comparison of the few-shot learning task in Fig.~\ref{fig:fewshot_workflow} by using existing formats and searched sparse format (denoted as ours). The settings of \textbf{M} and \textbf{m} is consistent with the columns in Fig.~\ref{fig:searched_index}. Ratio represents the image sampling ratio, which is the value of \textbf{n}/\textbf{N}.}
\label{tab:few_shot_comp}
\vspace{-5mm}
\end{table}

\begin{figure}[t]
\begin{minipage}[t]{\textwidth}
  \begin{minipage}{0.65\textwidth}
  \captionsetup{type=figure}
    \centering
    \subfloat[300W full: 68 landmks]
    {\includegraphics[width=0.475\textwidth]{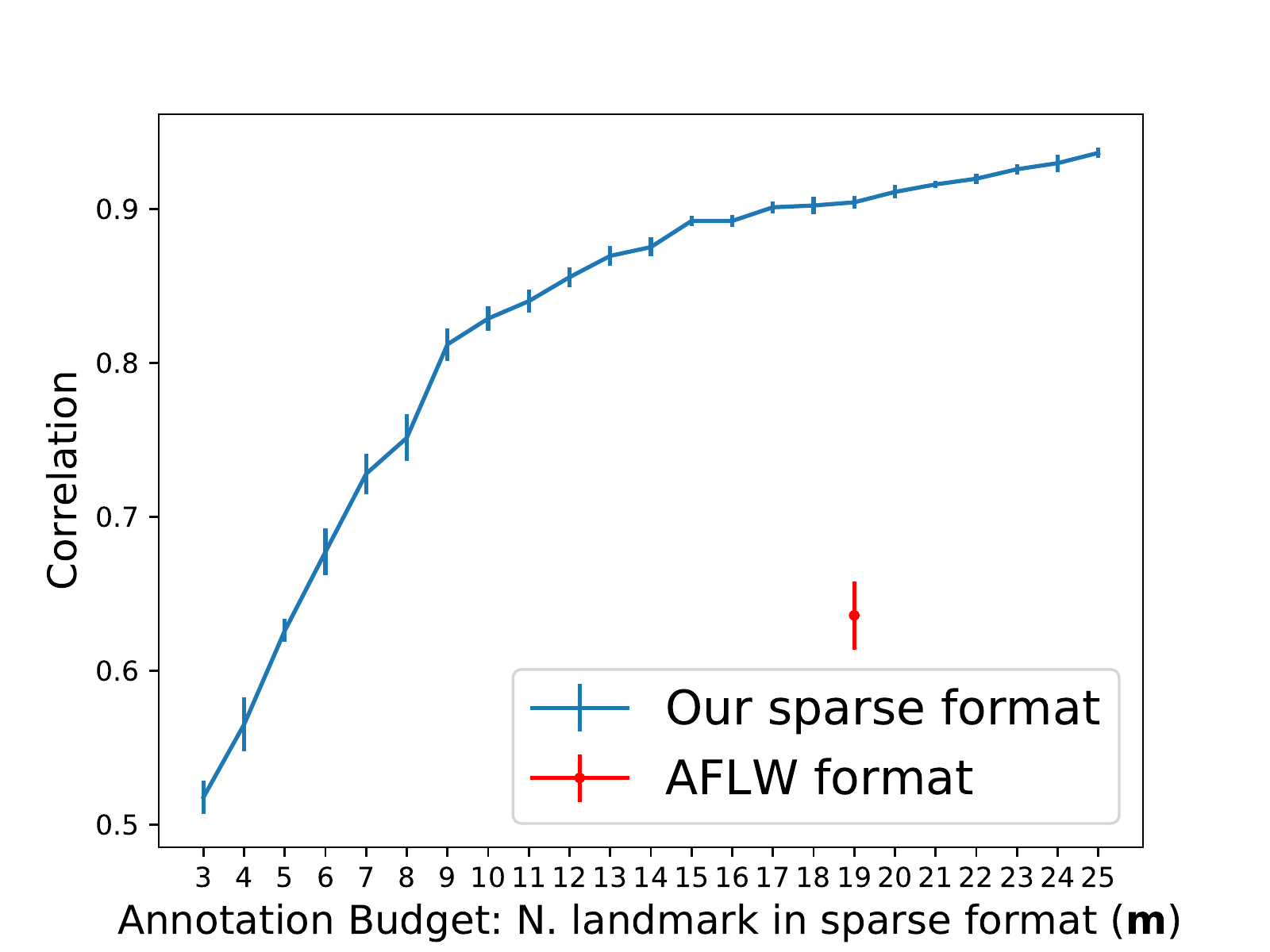}}
    \subfloat[WFLW: 98 landmks]
     {\includegraphics[width=0.475\textwidth]{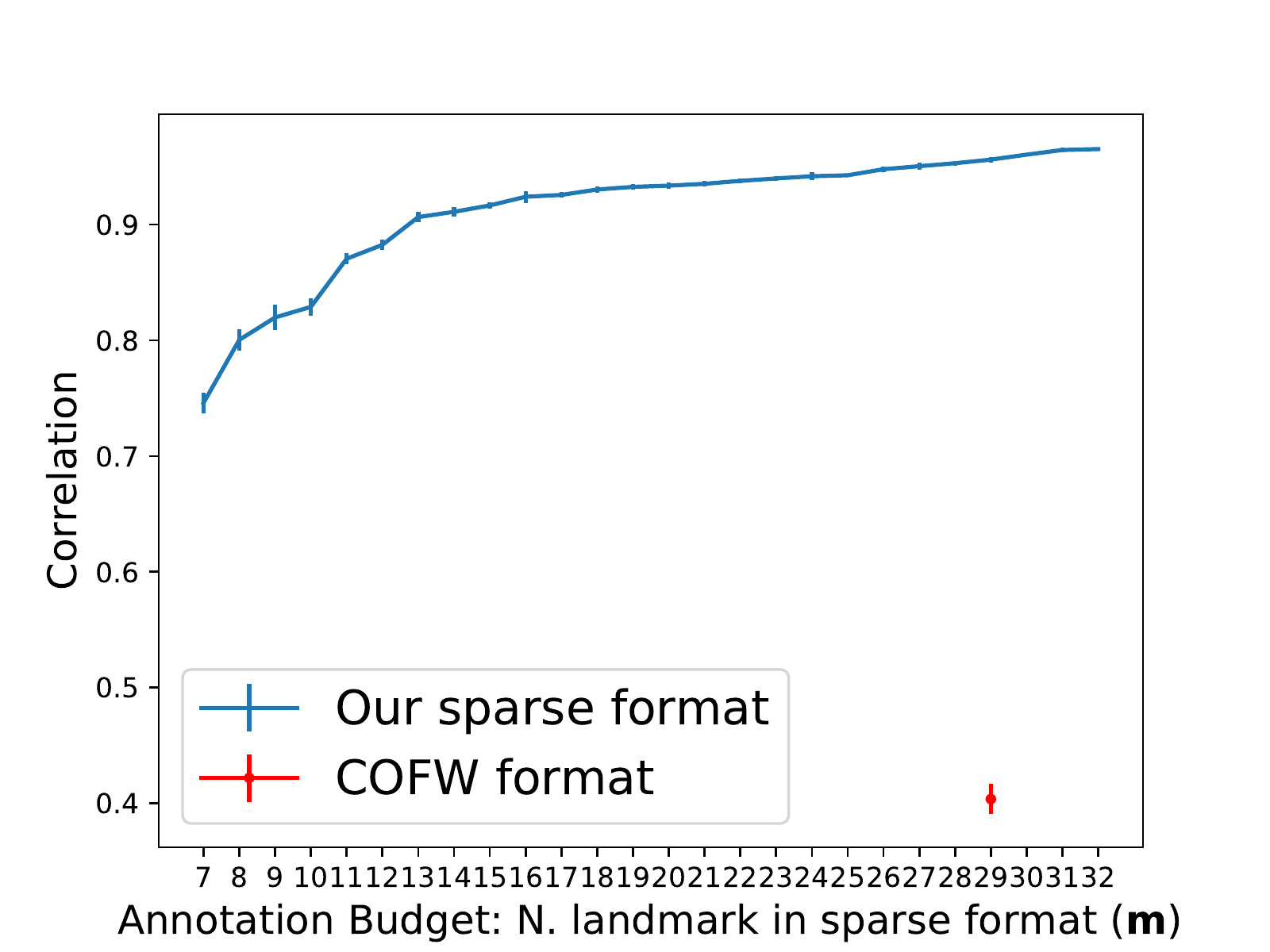}}
    \caption{Relationship between \textbf{m} and maximized minimum correlation $\mathbf{\hat{c}}$. For each \textbf{m}, we run searching method 10 times and plot the mean and variance.}
    \label{fig:search_curve} 
  \end{minipage}
  \hfill
  \begin{minipage}{0.34\textwidth}
    \centering
    \includegraphics[width=\textwidth, height=3.4cm]{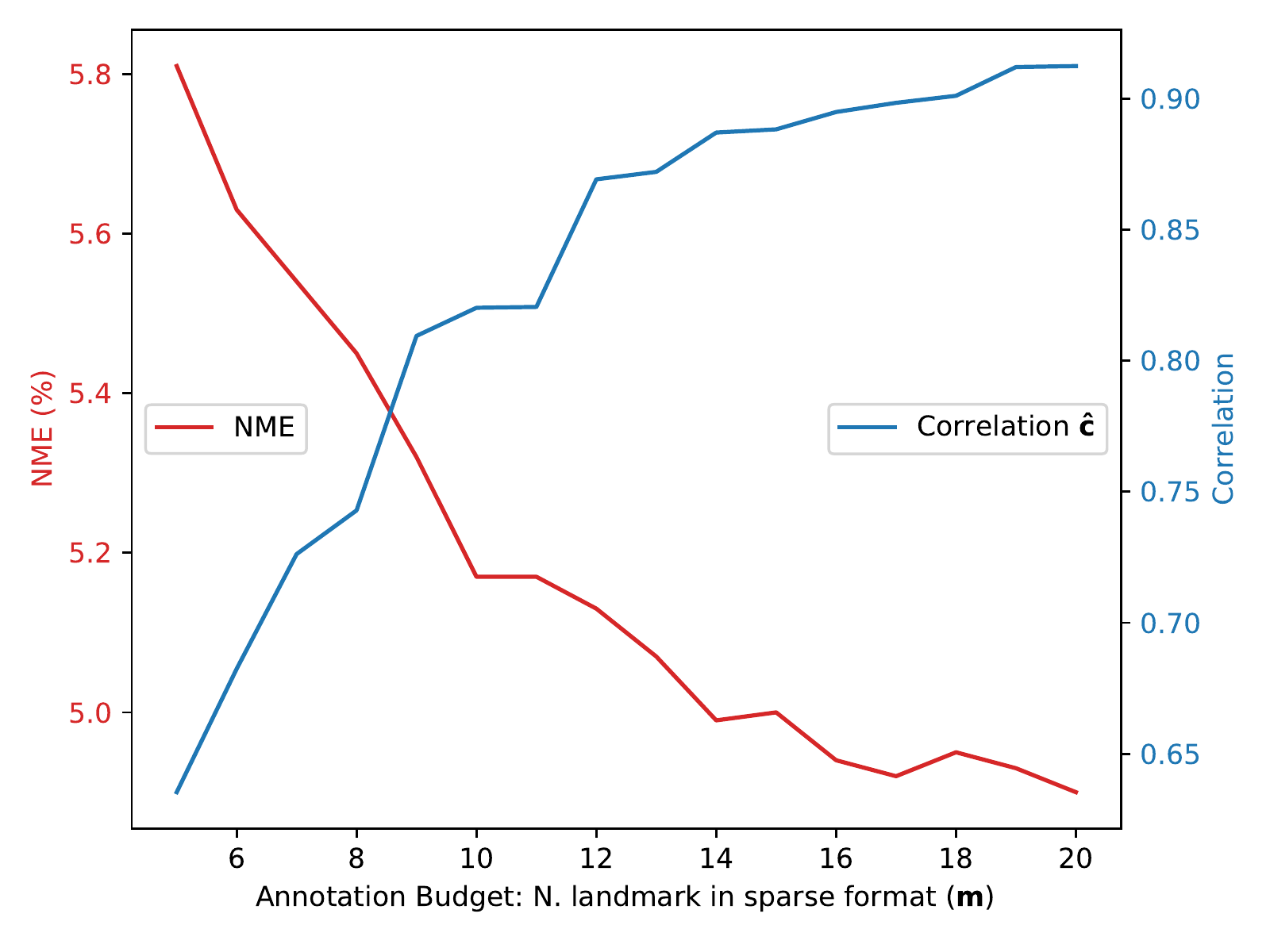}
    \captionof{figure}{Evolution of NME and $\mathbf{\hat{c}}$ with different \textbf{m} on 300W. \textbf{M}=68, \textbf{m}=3-20.}
    \label{fig:incremental_68}
    \end{minipage}
\end{minipage}
\end{figure}

To further investigate the relationship between the annotation budget~\textbf{m} and the maximized minimum correlation $\mathbf{\hat{c}}$ on different dense formats, we run our sparse format searching method on each dataset with incremental~\textbf{m}. The relationship between $\mathbf{\hat{c}}$ and \textbf{m} is shown in Fig.~\ref{fig:search_curve}. We also plot the values of $\mathbf{c}$ when using existing sparse formats. Our search method is able to find a bigger $\mathbf{c}$ compared to the existing formats, which can result in better performance on the few-shot learning task. This figure is useful for us to choose an appropriate annotation budget \textbf{m}. For example, on the 300W full format, we find a significant improvement on $\mathbf{\hat{c}}$ with 9 landmarks than 8 in the sparse format. It indicates that it is more advantageous to set the annotation budget \textbf{m} to 9 than 8 because the performance can probably be largely improved by adding only 1 landmark.\par

To validate the previous assumption that the performance of the few-shot learning task is strongly related to the maximized minimum correlation $\mathbf{\hat{c}}$. We plot the relationship between the performance of our few-shot learning tasks (NME \%) and $\mathbf{\hat{c}}$ with different \textbf{m} in our sparse format (see Fig.~\ref{fig:incremental_68}). We find that as $\mathbf{\hat{c}}$ goes up, the NME is improved accordingly, which confirms that the value of $\mathbf{\hat{c}}$ can be used as a guidance to choose a proper annotation budget \textbf{m}.

\section{Conclusions}

We propose a correlation analysis as a simple yet effective tool to interpret the relationship among facial landmarks. Our analysis provides a new perspective which is completely different to the commonly used metric NME. Conducting this analysis on the output prediction, we gain some interesting insights on the three most important models in the last decade.  
We also propose a few-shot learning method to drastically reduce the cost of laborious manual dense annotation. 
Our methodology on the coordinate correlation can be further extended to 3D facial landmarks, hand/body pose, object landmarks and even the bounding boxes of object detection. 
\par
%
%
\bibliographystyle{splncs04}
\bibliography{egbib}
\end{document}